\documentclass{article}
\usepackage{arxiv}
\usepackage[utf8]{inputenc} 
\usepackage[T1]{fontenc}    
\usepackage{hyperref}       
\usepackage{url}            
\usepackage{booktabs}       
\usepackage{amsfonts}       
\usepackage{nicefrac}       
\usepackage{microtype}      
\usepackage{lipsum}		
\usepackage{graphicx}
\usepackage{natbib}
\usepackage{doi}
\usepackage{tabularx}
\usepackage{subcaption}
\usepackage{array}
\usepackage{float}
\usepackage{algorithm, algorithmic}
\usepackage{amsmath, amssymb}
\usepackage[table]{xcolor}
\usepackage{multirow,multicol}
\usepackage{booktabs}   
\usepackage{makecell}   

\title{Colour Extraction Pipeline for \emph{Odonates} using Computer Vision}
\author{ \href{https://orcid.org/0009-0001-7357-1176}{\includegraphics[scale=0.06]{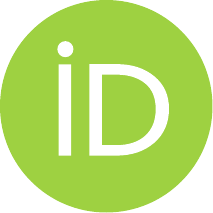}\hspace{1mm}Megan M.S. Rajaraman}\\
	Leiden Institute of Advanced Computer Science (LIACS)\\
	Leiden University\\
	Leiden, The Netherlands \\
	\texttt{mirnalinirms@gmail.com} \\
	\And
	\href{https://orcid.org/0000-0003-2445-8158}{\includegraphics[scale=0.06]{orcid.pdf}\hspace{1mm} Fons J. Verbeek} \\
	Leiden Institute of Advanced Computer Science (LIACS)\\
	Leiden University\\
	Leiden, The Netherlands \\
	\texttt{f.j.verbeek@liacs.leidenuniv.nl} \\
    	\And
	\href{https://orcid.org/0000-0002-1484-7865}{\includegraphics[scale=0.06]{orcid.pdf}\hspace{1mm}Vincent J. Kalkman} \\
    Naturalis Biodiversity Center \\
    Leiden, The Netherlands\\
	\texttt{vincent.kalkman@naturalis.nl} \\
    \And
	\href{https://orcid.org/0000-0002-2970-1180}{\includegraphics[scale=0.06]{orcid.pdf}\hspace{1mm}Rita Pucci} \\
    Leiden Institute of Advanced Computer Science (LIACS)\\
	Leiden University\\
	Leiden, The Netherlands \\
	\texttt{r.pucci@liacs.leidenuniv.nl} \\
}

\date{}

\renewcommand{\shorttitle}{\textit{Colour Extraction Pipeline for Odonates}}

\hypersetup{
pdftitle={Colour Extraction Pipeline for Odonates using Computer Vision},
pdfsubject={q-bio.NC, q-bio.QM},
pdfauthor={David S.~Hippocampus, Elias D.~Striatum},
pdfkeywords={First keyword, Second keyword, More},
}

\begin{document}
\maketitle

\begin{abstract}
The correlation between insect morphological traits and climate has been documented in physiological studies, but such studies remain limited by the time-consuming nature of the data analysis. In particular, the open source datasets often lack annotations of species' morphological traits, making dedicated annotations campaigns necessary; these efforts are typically local in scale and costly. In this paper, we propose a pipeline to identify and segment body parts of Odonates (dragonflies and damselflies) using deep neural networks, with the ultimate goal of extracting body parts' colouration. The pipeline is trained on a limited annotated dataset and refined with pseudo supervised data. We show that, by using open source images from citizen science platforms, our approach can segment each visible subject (Odonates) into head, thorax, abdomen, and wings and then extract a colour palette for each body part. This will enable large-scale statistical analysis of ecological correlations (e.g., between colouration and climate change, habitat loss, or geolocation) which are crucial for quantifying and assessing ecosystem biodiversity status.

Code available at: https://github.com/itismeganrms/colour-extraction-odonates.git 

\end{abstract}

\section{Introduction}

 
 
\textit{Odonata} is a small order of predatory insects that are found ubiquitously, and are easy to spot. It comprises a few subspecies that include dragonflies and damselflies. They are characterized by compound eyes (made up of thousands of ommatidia) \cite{bybee_odonata_2016}, two pairs of strong wings and an elongated body. Dragonflies
are closely related to damselflies, and are very similar in appearance. 
They feed on a variety of insects ranging from mosquitoes to flies \cite{priyadarshana_meta-analysis_2023}. Their presence and feeding habits have greatly influenced both aquatic and terrestrial ecosystems, and are good indicators of the quality of aquatic habitats \cite{may_odonata_2019}. They also exhibit colour polymorphism \footnote{Colour polymorphism refers to the existence of two or more discrete colour phenotypes within the same population. This allows the males to `hide' from fighter males, by camouflaging, while still being near females, and the females to avoid harassment \cite{gossum_evolution_2008}. } This behaviour is greatly influenced and affected by the temperature of the location. Higher temperatures affect wing colouration, which affects the flight and performance of the males \cite{moore_temperature_2019}. In addition to this, there is also evidence to suggest that the colour changes with respect to the latitude \cite{hassall_effects_2008}. As they are tropical in nature, temperature changes affect their physiology greatly, namely their developmental rate, immune function and the development of pigment for thermoregulation \cite{hassall_effects_2008}. This, in turn, affects the flight and movement of Odonates. As they are predatory insects, this could also negatively influence the ecosystem by disturbing the natural balance of insect species. Such creatures are now being threatened due to habitat destruction, and clearance of forests \cite{samways_scientists_2025}. Hence, there exists an urgent need to understand the change in colouration and influence of environmental and geographical factors, on the colouration. To the best of our knowledge, there are no readily available datasets that present the colour information with annotation of the Odonates. 

To achieve this on a global scale, computer vision models are necessary to identify and segment the Odonates. Manual monitoring of insects is laborious and time-intensive, as they involve setting up insect fly traps and using the captured specimens for the creation and curation of the dataset. However, they often capture not only the species of interest, but other insects as well. These insects need to be photographed and annotated in order to train specific models \cite{jain_insect_2024}. Automated identification of insects is also challenging, as the insects move quickly and are small in size. They are often occluded by flowers or leaves. As a result, models often struggle with separating the object of interest and the background \cite{bjerge_motion_2023}. Therefore, there exists a need for models that are trained specifically on Odonates and on readily available data. This would allow us to build a generalized pipeline that can be used and extended easily, without prior knowledge. 

The approaches listed in this paper aim to address some of the aforementioned gaps by utilizing data from citizen science publicly available online. This paper proposes a new computer vision-based pipeline to support the extraction of colour from Odonates, and focuses on analysis of body colours and the differences in colours among the body parts of the insects. The models used for this paper are built and trained on citizen science data. The metadata from the images provides the information required for geo-specific and ecological correlation analyses. As the datasets are not prepared to facilitate the segmentation of the parts of the body for colour extraction, one of the tasks for the project is the annotation of a functional dataset, for instance and semantic segmentation. As there are no available datasets that contain the Odonates and the required parts, a significant part of this paper focused on manual annotation to train the models. A small portion of the dataset was manually annotated and used for the first round of experiments and fine-tuning. The results of the first round of fine-tuning experiments provided a well-performing model, which was used for additional annotation. The second round of fine-tuning experiments was done on a combination of the datasets after both rounds of annotation. The best-performing model after the second round of experiments was chosen as the final model, which was used for identifying and segmenting the parts of the Odonate (head, thorax, abdomen, and wings) from the image. The final task of this project focuses on providing a preliminary exploratory analysis and an initial pipeline for the extraction of colour from each identified part, and enabling statistical analysis of ecological correlations between colour distribution and geolocalization, as well as between colour distribution and time of the day. 

Similar approaches have been done for annotation and identification of multiple classes of insects, such as this paper \cite{orsholm_multi-modal_2025}, which proposes a large-scale insect dataset MassID45, encompassing 17 species and 35,586 images. This paper focuses solely on annotation, instance and semantic segmentation of these species. In another paper \cite{idec_using_2024}, the authors implement a similar statistical analysis which focuses on global geography of ant colour. Such segmentation or colour extraction pipelines have either been implemented for multiple insect classes, or focuses on a different species other than Odonates. This illustrates the need for a specialised study on Odonates. 

The paper is therefore be divided into three main cores: annotation and preparation of the dataset, instance and semantic segmentation of the object and extraction of colour from said object. All subsequent chapters will be structured similarly for easier understanding. 

\section{Related Works} \label{sec:headings}
\paragraph{Annotation}



For instance, for semantic segmentation of the objects of interest, accurate annotation of the objects of interest is essential. Benchmark datasets used for semantic and instance segmentation, such as Microsoft COCO \cite{lin_microsoft_2015}, Cityscapes Dataset \cite{cordts_cityscapes_2016}, and Mapillary Vistas Dataset \cite{neuhold_mapillary_2017}, were annotated with the help of in-house annotators and quality control tools or outsourced to third-party platforms. This paper focuses on manual annotation, as automated annotation did not yield reliable results, and manual annotation allowed us to accurately define the objects of interest and boundaries.

We observed annotation pipelines in other research fields to understand how this task was tackled. Common manual annotating tools use a polygon-based approach, which does not seem to provide the granularity that is required for capturing the wing curvature, or the thorax of the Odonates. In histology, a commonly used tool is QuPath \cite{bankhead_qupath_2017}, which is open-source, and has tools like the Magic Wand, which allows brushing over the parts with precision, and even a PyTorch plugin to run models for annotation. This tool provided the required flexibility, and hence, QuPath was chosen for the first phase of manual annotation. 

Another common approach is annotating a small portion of the dataset and using models or external tools to annotate larger portions of the dataset, such as MassID45 \cite{orsholm_multi-modal_2025}, which was the approach chosen for the second phase of the project. 

\paragraph{Segmentation Models}
Semantic and instance segmentation are different segmentation tasks, which help in recognizing different classes of objects and different instances of the same class, respectively. This paper deals with both instance and semantic segmentation, as it requires the identification of the Odonates and isolation of the different parts.

Despite recent advancements in semantic and instance segmentation, very few papers and current implementations focus on insects. Some focus on animal such as \textit{Fantastic Animals and Where to Find them} \cite{zhang_fantastic_2024}, which explores segmentation for marine animals using DualSAM and deals with the occlusion and lighting that are typically found in marine images. Another paper is \textit{Learning Part Segmentation from Synthetic Animals} \cite{peng_learning_2023}, which deals with semantic part segmentation in synthetic animals, and uses Skinned Multi-Animal Linear (SMAL) models for segmentation. Some articles examine insect segmentation, such as this paper \cite{kargar_tiny_2025}, which implements a U-Net model to count and segment the stink bug, and the paper discussed previously (MassID45) \cite{orsholm_multi-modal_2025}, which looks at the identification of multiple classes of insects. 

\paragraph{Colour Extraction}
Colour models like RGB, HSV and CIELAB \footnote{RGB model is an additive model for light where the colours are represented in primary pigments of \textbf{R}ed, \textbf{B}lue and \textbf{G}reen and additive combinations of the colours. Similarly, HSV is a colour model which depicts the \textbf{H}ue, \textbf{S}aturation and \textbf{V}alue of an object. CIELAB, also known as L*a*b* is used to denote colours in the relationship of visible spectrum of light and human vision. The L* denotes the relative lightness, and a* and b* refer to the primary four colours observed: red, green, blue, and yellow. } are commonly used to understand the surrounding colour.
While these colour models exist for representation and understanding, colour extraction and analysis is usually done in the hyperspectral space. This has been done in multiple fields, such as astronomy or in the medical domain. Even within insects, hyperspectral imaging has been the primary tool of choice \cite{foster_hyperspectral_2019} to identify different species 
\cite{wang_rapid_2024}, \cite{tan_leveraging_2024} or analyse development \cite{lacotte_comparative_2023}. However, hyperspectral imaging is laborious, and requires an expensive hyperspectral camera as well as specimens to record the samples. As the goal of the project was to devise a method that works for large-scale, readily available datasets, and it is not feasible to record 300,000 specimens, another method was implemented. 


There are a few existing papers that look at colour analysis and extraction on RGB or HSV/HSI scale. Such papers are on the green anole (\textit{Anolis carolinensis}), which uses K-Means Clustering to cluster the dominant colours in RGB space \cite{price_using_2025} and on ant colour \cite{idec_using_2024}, which extracts the dominant HSV values for each species, and looks at the variation over the climate and location. Both these papers proved instrumental for the colour extraction and analysis of this project. 

\section{Methodologies}
\subsection{Dataset Acquisition and Manual Annotation of the Objects}
As the dataset was created and curated from citizen science data, there were no readily available ground truth labels. Furthermore, when an unseen image was used for inference on models used for annotation, it was observed that the models struggled to identify the parts accurately (the results of which are presented in \autoref{sec:zero-shot-learning}). Hence, annotations were done manually for accurate identification. 
\subsubsection{Dataset Acquisition}
The dataset comprises insects of the order \texttt{Odonata} observed in Europe. In order to extract the colour from the Odonates, only adults i.e., \texttt{Imago} life stage were considered. This also ensures that the wings are functional and fully developed. After filtering, the dataset consists of 759,423 records from 73 published datasets. The DOI of the dataset is given here \cite{gbiforg_user_occurrence_2025}. 

When doing a manual check of some of the images in the dataset, it was observed that some of the images were blurry, or the subject was well-camouflaged in the leaves or the surroundings. As they are citizen science image, there were many images where the hands occluded parts of the Odonate. Hence, these images were removed from the dataset. 


\subsection{Annotation}
After initial analysis and pruning of the dataset, 70 random images were selected for manual annotation. Annotation was done for the head, thorax, abdomen, and wings of the Odonate. Some manually annotated images were checked by an expert entomologist to ensure that the parts are identified accurately, and the boundaries between each part are captured accurately. 



For the first round of experiments, annotation was done using QuPath \cite{bankhead_qupath_2017} (explained in detail in Section~\ref{sec:qupath}) and converted to YOLO-compatible format using Python code. However, the other three models use COCO-style annotations \footnote{YOLO uses annotations structured as images and labels in the form of text files, and arranged in a specific folder format. Each line in the text file corresponds to an annotated object in the image and is written as a string of numbers which contain the  \texttt{class x\_center y\_center width height}. However, COCO-style annotations consist of all the images and annotations structured as a JSON file, with the class ID for each annotation at the end. The aforementioned structure is used only by YOLO, whereas COCO-style annotations are used by models that are built on Detectron2 library.}
Hence, the annotations were uploaded and validated again using RoboFlow \cite{dwyer2025roboflow} and downloaded in both YOLO and COCO formats. The first version of the dataset is linked here \cite{odonata-v1}. 

For the second round of experiments, as the annotations were generated from the trained YOLO-exp01 model, an extension called YOLO-Label \cite{noauthor_andaoaiyolo-label-vs_nodate} was used to validate the generated annotations and make modifications where the model erred. 
Again, we used RoboFlow \cite{dwyer2025roboflow} to validate and export the annotations in the supported formats for the models. The second version of the dataset is linked here \cite{odonata-v2}. 

\subsubsection{QuPath} \label{sec:qupath}
\begin{figure}[H]
    \centering
    \begin{subfigure}[b]{0.49\linewidth}
        \centering
        \includegraphics[width=\linewidth]{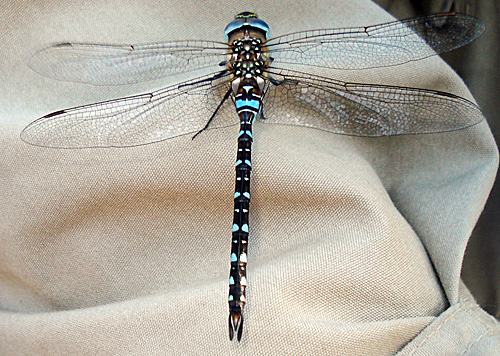}
        \caption{Original Image}
    \end{subfigure}
        \begin{subfigure}[b]{0.49\linewidth}
        \centering
        \includegraphics[width=\linewidth]{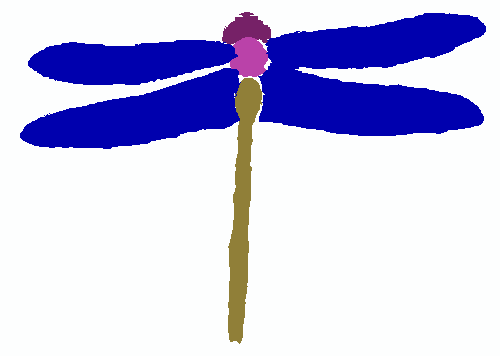}
        \caption{Annotated Mask}
    \end{subfigure}
    \caption{Manual annotation of dragonfly and resulting segmentation mask from QuPath. The first image is the original image and the second image is the annotated image, which has been exported as .TIFF file}
    \label{fig:segmentation-example}
\end{figure}

QuPath is an open-source annotation tool primarily used for bio-image analysis, such as histological and pathological data. The Wand tool provided in QuPath helped draw accurate annotations for the wings and helped capture the contours of the wings accurately. In this manner, 70 images were manually annotated, and the labels were exported as TIFF images, as seen in \autoref{fig:segmentation-example}. 
The exported TIFF files were then used to create YOLO style annotations for the first round of experiments.    
\subsection{Model Architectures}
Four models were chosen for training: two CNN-based models and two transformer-based models. These four models were chosen to observe how different architectures recognize, and to test four state-of-the-art models that are commonly used for segmentation tasks, on the dataset. 
\begin{enumerate}
    \item YOLOv11 (YOLOv11x-seg) \cite{khanam_yolov11_2024}: This model follows a CNN based architecture, and is one of the state-of-the-art models for image segmentation. The model processes images as a whole, so as to capture all the contextual information. 
    \item Mask R-CNN \cite{he_mask_2018}: This model also follows a CNN based architecture, and is an extension of Faster R-CNN. This utilizes ROIAlign and another mask head for generating segmentation masks. 
    \item MaskDINO \cite{li_mask_2022}: This model is based on a transformer architecture, and is an extension of DINO. It uses content-query embeddings and a mask head (similar to the above-mentioned model) for predicting masks.
    \item Mask2Former \cite{cheng_masked-attention_2022}: This model also follows a transformer-based architecture, and is an extension of MaskFormer. It utilizes two mask heads, one for embeddings and the other for extracting localized features. 
\end{enumerate}
\subsection{Colour Extraction} \label{sec:hypotheses-colour-extraction}
The final phase of the pipeline consists of colour extraction. The colour extraction was done in two ways: using K-Means Clustering to get the dominant colour, and by using colour models to obtain the mean values of each part. The species chosen for this part of the paper is \textit{Sympetrum striolatum}, which is predominantly found in the Netherlands (and along the borders of Belgium and Germany). This species was chosen, as it exhibits colour polymorphism, and also changes colour with respect to geographic location or hour of day. 
\subsubsection{K-Means Clustering} \label{sec:kmeans-methods}
The dominant colour and hue is determined by K-Means Clustering. As observed in the below equation, the objective is dividing the observations into sets or clusters.  
\begin{equation*}
     \underset{c}{\text{arg}\text{min}} \sum_{i=1}^{k} \sum_{x \in S_i} \left\| x - \mu_i\right\|^2 =  \underset{c}{\text{arg}\text{min}} \sum_{i=1}^{k} |S_i| \text{Var} S_i
\end{equation*}
Here, $S_i$ refers to the dominant hue, and $k$ refers to the desired number of clusters. In the context of the current pipeline, the best model was used to generate prediction masks for the image. The masks of the head, thorax, and abdomen are considered for the colour extraction and analysis, as wings are often transparent or iridescent. This would affect the analysis and not portray an accurate representation, as the transparent wings often show the scene under the wings. 

Using the predicted masks, the colour extraction is done as explained in \autoref{alg:kmeans-hue} below. 
\begin{algorithm}[H]
\begin{algorithmic}[1]
\STATE Obtain the class ID and masks for each part (head, thorax, and abdomen) of the Odonate. \\
\FOR {the class ID and corresponding mask}
\STATE resize the mask of the identified part to the original image\\
\STATE Grey Scale Mask $\leftarrow$ Convert the mask into a greyscale mask by multiplying with 255 \\
\STATE Binary Mask $\leftarrow $Thresholding the grey scale mask \\
\STATE Final part in colour is obtained by $\text{image} \land \text{image}$ and using the binary mask\\
\ENDFOR
\STATE Use K-means clustering to obtain n- number cluster of colours from the final part \COMMENT{n=5 in this case}\\
\STATE Obtain final histogram of colours and frequency of occurrence
\end{algorithmic}
\caption{Construction of palette using K-Means Clustering (taken from \cite{noauthor_dominant_nodate})}
\label{alg:kmeans-hue}
\end{algorithm}

The final result of the algorithm is a histogram of colours ordered by the occurrence. This acts as the palette of colours for each part of the body. 

\subsubsection{Extraction of HSV values} \label{sec:hsv-methods}
For the second part of colour extraction and statistical analysis, a similar yet different approach was taken. 

This approach was inspired by the statistical analysis done on ants \cite{idec_using_2024}. As implemented there, the average HSV values of each part are extracted, and used for statistical analysis. To observe that, the mean lightness (\textbf{V}) value is taken, as it indicates how light or dark the colour is. This allows to observe if the level changes with respect to latitude or hour of day. 

The first part of the extraction follows the same method, as shown in \autoref{alg:kmeans-hue}. However, after identifying the parts of the subject, the average RGB and HSV values are extracted for analysis. This is done with the help of \texttt{cvtColour} function from the python package \texttt{OpenCV}. 

\section{Experiments} \label{sec:experiments}
\begin{figure}[H]
    \centering
    \includegraphics[width=1\linewidth]{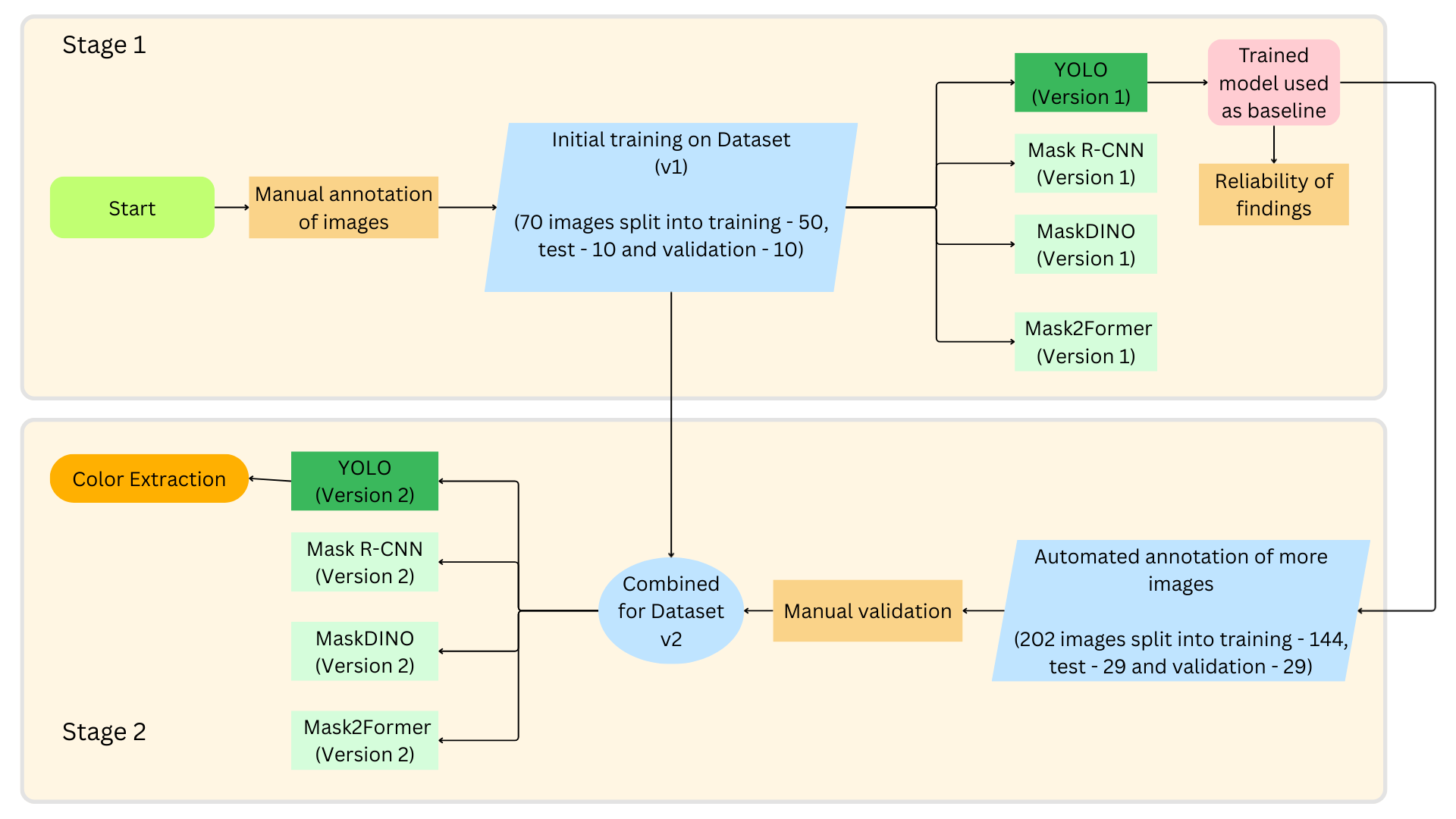}
    \caption{A flowchart representing the research pipeline for the project}
    \label{fig:research-flowchart}
\end{figure}

As show in \autoref{fig:research-flowchart}, the research pipeline is organized in two main stages. 
\paragraph{Stage 1} \label{sec:experiments-stage1}
The project starts with 70 images which were manually annotated on QuPath. Using the 70 images, the four selected models were initially trained to check the performance and gaps for further refinement. At this stage, the best-performing model is selected (as highlighted by the darker green highlight) as discussed in Section~\ref{sec:fine-tuning-exp-01}. The trained model is used as a temporary baseline in Stage 2, while the reliability of the best-performing model is tested and validated by a survey and ratings from experts (this has been discussed in detail in \autoref{sec:survey-results}). 
\paragraph{Stage 2} \label{sec:experiments-stage2}
As the dataset was not sufficient for further experiments, the best-performing model (YOLO) was used to generate predictions on more images, so that it can be used for further training.  The model was used to generate annotations for more images (202 more). Manual validation of the images were done to improve the quality of automatic annotation. The final corrected set of images was combined with the first set of images to form the second version of the dataset. The second version of the dataset was used to improve the performance of YOLO, and the other three models. 

In this stage, the second task of colour extraction was also addressed. Extraction of colour was done by two ways, as explained in \autoref{sec:hypotheses-colour-extraction}. For extracting the colour and general statistical analysis of the dataset, metadata such as the latitude, longitude, timestamp of capture, general location and gender were utilized.

\subsection{Model Training and Performance}
The four models have been trained on two versions of the dataset, as mentioned in the earlier section.

\begin{table}[H]
    \centering
    \begin{tabular}{c>{\centering\arraybackslash}p{2cm}>{\centering\arraybackslash}p{2cm}}
        \hline
        \multirow{2}{*}{\textbf{Split}} & \multicolumn{2}{c}{\textbf{Count of Samples}} \\ \cline{2-3}
         & \textbf{v1} & \textbf{v2} \\ \hline
      Training & 50 & 194 \\ \hline
      Validation&  10 & 39\\ \hline
      Test&10 & 39 \\ \hline
    \end{tabular}
    \caption{Dataset Split and Count per split}
    \label{tab:dataset-split}
\end{table}

The first model (YOLO) has been loaded directly from Ultralytics, and is pretrained on ImageNet. The other three models (Mask R-CNN, Mask2Former, MaskDINO) are all forks of Detectron2, and pretrained on MS-COCO.
\begin{enumerate}
    \item \textbf{Fine-tuning: Stage 1} All four models that have been pretrained on large-scale datasets, are trained on v1 of the dataset, and the performance has been recorded.
    \item \textbf{Fine-tuning: Stage 2}All four models that have been pretrained on large-scale datasets, are trained on v2 of the dataset, and the performance has been recorded.
\end{enumerate}

\section{Results}
The results of this paper are divided into two sections: results based on the segmentation task, and results based on the colour extraction. 
\subsection{Results based on Segmentation Task}
The results of the models are evaluated based on two metrics, namely the mean average precision (mAP) and the average precision (AP). The mAP is taken along with an IoU threshold of 0.5 and 0.75, while the AP values are calculated for each class. These thresholds are chosen, as they are typical benchmark metrics. 
\subsubsection{Fine-tuning : Stage 1} \label{sec:fine-tuning-exp-01}
As explained in \autoref{sec:experiments-stage1}, the first experiment consists of observing the performance of the models on the first version of the dataset \cite{odonata-v1} (70 images). The results of the first experiment are divided into two tables: \autoref{tab:Initial-seg-results-bbox}, which records the performance of the model in determining the bounding boxes for the parts of the body, i.e.,  the identification of the Odonate, and \autoref{tab:Initial-seg-results-seg}, which records the performance of the model on the generation of masks for the parts.

YOLO-exp01 was trained for 150 epochs, while we set up training for the other three models to 100, 200 and 600 epochs. \footnote{YOLO-exp01 was trained for 150 epochs, as extensive hyperparameter tuning was done for this model (the results of which are not included in this paper) and 150 epochs was observed as the point of convergence. }
\begin{table}[H]
\centering
\normalsize   
\begin{tabular}{ccccccccc}
\toprule
    \multirow{2}{*}{\textbf{Model}} & \multirow{2}{*}{\textbf{Epochs}} & \multicolumn{3}{c}{\textbf{Mean Values}} & \multicolumn{4}{c}{\textbf{Per-Class Values}} \\\cmidrule(lr){3-5} \cmidrule(lr){6-9} 
    & & \textbf{mAP} & \textbf{mAP50} & \textbf{mAP75} & \textbf{AP-head}  & \textbf{AP-thorax} & \textbf{AP-Abdomen} & \textbf{AP-wings}  \\ \midrule
    \rowcolor{lightgray}\textbf{YOLO-exp01} & \textbf{150} & \textbf{50.465} & \textbf{79.802} & \textbf{55.606} & \textbf{61.093} & \textbf{43.515} & \textbf{50.697}  & \textbf{46.554} \\ \midrule
    
    \multirow{3}{*}{Mask R-CNN} & 100 & 8.83 & 24.952 & 6.188 & 17.822 & 5.183 & 12.315 & 0 \\ \cmidrule(lr) {2-9} 
                                & 200 & 8.666 & 26.603 & 5.507 & 16.747 & 11.868 & 6.048 & 0 \\ \cmidrule(lr) {2-9} 
                                & 600 & 7.513 & 27.042 & 2.547 & 16.457 & 6.671 & 6.922 & 0 \\\midrule
                                 
    \multirow{3}{*}{Mask DINO} & 100 & 0.045 & 0.454 & 0 & 0.182 &0 &0 &0 \\ \cmidrule(lr) {2-9} 
                               & 200 & 4.808 & 9.385 & 3.837& 17.871& 1.361&0 &0 \\\cmidrule(lr) {2-9} 
                               & 600 & 0.472 & 1.906 & 0.297 & 0.338 &0.203& 1.347 & 0 \\ \midrule
                                
    \multirow{3}{*}{Mask2Former} & 100 & 0 & 0 & 0 & 0 & 0 & 0 & 0 \\ \cmidrule(lr) {2-9} 
                                 & 200 & 0 & 0 & 0 & 0 & 0 & 0 & 0 \\ \cmidrule(lr) {2-9} 
                                 & 600 & 0 & 0 & 0 & 0 & 0 & 0 & 0 \\ 
\bottomrule
\end{tabular}
\caption{Performance of the model is documented for the \textbf{bounding boxes}, i.e., identification of the \textit{Odonate}, as part of Stage 1 experiments. Training and testing was done on the first version of the dataset \cite{odonata-v1}, as discussed in \autoref{sec:experiments-stage1}. The metrics are recorded for the mean average precisions and average precision for each class, and the best performing model is in bold. }
\label{tab:Initial-seg-results-bbox}
\end{table}
\begin{table}[H]
\centering
\normalsize   
\begin{tabular}{ccccccccc}
\toprule
    \multirow{2}{*}{\textbf{Model}} & \multirow{2}{*}{\textbf{Epochs}} & \multicolumn{3}{c}{\textbf{Mean Values}} & \multicolumn{4}{c}{\textbf{Per-Class Values}} \\\cmidrule(lr){3-5} \cmidrule(lr){6-9} 
    & & \textbf{mAP} & \textbf{mAP50} & \textbf{mAP75} & \textbf{AP-head}  & \textbf{AP-thorax} & \textbf{AP-Abdomen} & \textbf{AP-wings}  \\ \midrule
    \rowcolor{lightgray}\textbf{YOLO-exp01} & \textbf{150} & \textbf{40.213} & \textbf{81.614} & \textbf{35.420} & \textbf{45.915} & \textbf{36.961} & \textbf{39.862}  & \textbf{38.116} \\ \midrule
    
    \multirow{3}{*}{Mask R-CNN}  & 100 & 1.458 & 9.752 &0 & 5 & 0.832 & 0 & 0 \\ \cmidrule(lr) {2-9} 
                                 & 200 & 8.666 & 26.603 & 5.507 & 16.747 & 11.868 & 6.048 & 0 \\ \cmidrule(lr) {2-9} 
                                 & 600 & 1.337 & 6.58 & 0 & 1.312 & 0.545 & 3.49 &0 \\ \midrule
    \multirow{3}{*}{Mask DINO}   & 100 & 0.227 & 0.454 &0 &0.908 &0 &0 &0 \\\cmidrule(lr) {2-9} 
                                 & 200 & 5.124 & 10.190& 2.867 & 18.218 & 2.277 &0 &0 \\ \cmidrule(lr) {2-9} 
                                 & 600 & 0.799 & 1.773 &0.421 &0.297 &0.657 &2.243 &0 \\ \midrule
    \multirow{3}{*}{Mask2Former} & 100 & 0& 0 &0 &0 & 0&0 & 0  \\\cmidrule(lr) {2-9} 
                                 & 200 & 0.087 & 0.693& 0&0 &0 &0.35 & 0 \\ \cmidrule(lr) {2-9} 
                                 & 600 & 0.816 & 2.846 & 0 & 0.581 &0  & 2.682 & 0 \\ \bottomrule
\end{tabular}
\caption{Performance of the model is documented for the \textbf{masks} i.e., segmentation of the parts of the \textit{Odonate}, as part of Stage 1 experiments. Training and testing was done on the first version of the dataset \cite{odonata-v1}, as discussed in \autoref{sec:experiments-stage1}. The metrics are recorded for the mean average precisions and average precision for each class, and the best performing model is in bold. }
\label{tab:Initial-seg-results-seg}
\end{table}

Based on the results, it is observed that YOLO-exp01 outperformed the other three models, and was the only model out of the four to classify and segment the wings. Additional inference was run on a batch of unseen images, and it was observed that YOLO-exp01 was able to accurately identify and segment all four parts. However, it struggled with drawing a correct boundary between the thorax and abdomen. Mask R-CNN was able to classify and segment the head, thorax, and abdomen of the Odonate, but could not identify the wings. MaskDINO was able to produce masks for each part, but struggled with identification and mislabelled the parts of the Odonate. Mask2Former struggled with both classification and segmentation, and produced partial masks for the parts. The inference of YOLO-exp01 on an unseen image is provided in \autoref{fig:finetune-yolo-1}. 

\subsubsection{Fine-tuning : Stage 2} \label{sec:fine-tuning-exp-02}
After the first round of experiments, the models were trained again on the second version of the dataset \cite{odonata-v2}. All four models were trained for 150 epochs, and the results are shown in \autoref{tab:refined-seg-results-bbox} and \autoref{tab:refined-seg-results-seg}. 
\begin{table}[H]
\centering
\normalsize   
\begin{tabular}{ccccccccc}
\toprule
    \multirow{2}{*}{\textbf{Model}} & \multirow{2}{*}{\textbf{Epochs}} & \multicolumn{3}{c}{\textbf{Mean Values}} & \multicolumn{4}{c}{\textbf{Per-Class Values}} \\\cmidrule(lr){3-5} \cmidrule(lr){6-9} 
    & & \textbf{mAP} & \textbf{mAP50} & \textbf{mAP75} & \textbf{AP-head}  & \textbf{AP-thorax} & \textbf{AP-Abdomen} & \textbf{AP-wings}  \\ \midrule
    \rowcolor{lightgray}\textbf{YOLO-exp02} & \textbf{150} & \textbf{64.667} & \textbf{91.859} &\textbf{ 70.381} & \textbf{67.946} & \textbf{53.991} & \textbf{67.162}  & \textbf{69.569} \\ \midrule
    Mask R-CNN & 150  & 21.368 & 46.229 & 15.414 & 39.868 & 29.447 & 16.155 & 0 \\ \midrule
    Mask DINO & 150 & 0.317 & 0.882 & 0.155 & 0.303 &0.027 &0.939 &0 \\ \midrule
    Mask2Former & 150 & 0 & 0 & 0 &0 & 0 & 0 & 0 \\\bottomrule
\end{tabular}
\caption{Performance of the model is documented for the \textbf{bounding boxes} i.e., identification of the \textit{Odonate}, as part of Stage 2 experiments. Training and testing was done on the second version of the dataset \cite{odonata-v2}, as discussed in \autoref{sec:fine-tuning-exp-02}. The metrics are recorded for the mean average precisions and average precision for each class, and the best performing model is highlighted in grey. }
\label{tab:refined-seg-results-bbox}
\end{table}

\begin{figure}[H]
    \centering
    \begin{subfigure}[b]{0.49\linewidth}
        \centering
        \includegraphics[width=\linewidth]{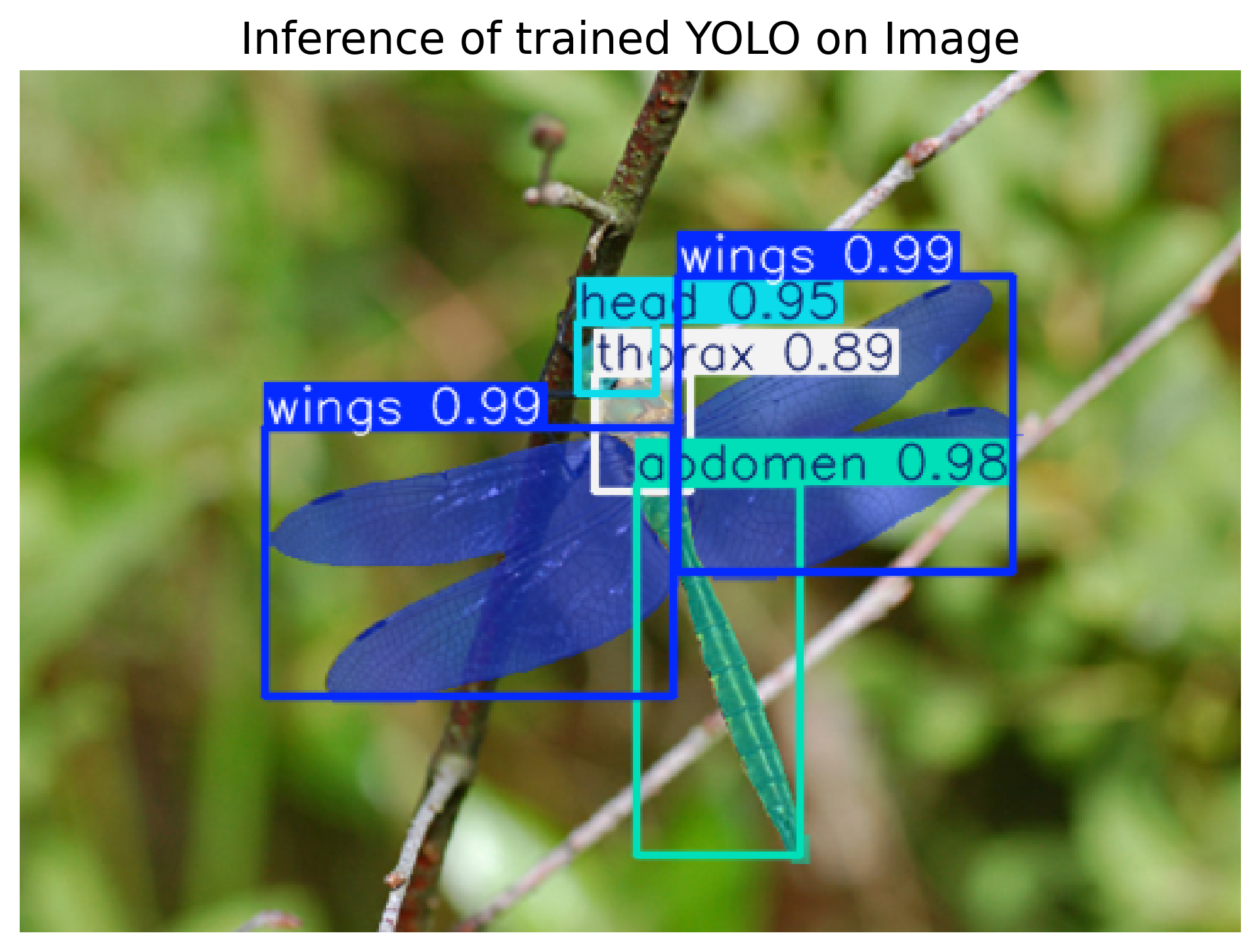}
        \caption{Fine-tuning Stage 1: Results of YOLO-exp01}
        \label{fig:finetune-yolo-1}
    \end{subfigure}
        \begin{subfigure}[b]{0.49\linewidth}
        \centering
        \includegraphics[width=\linewidth]{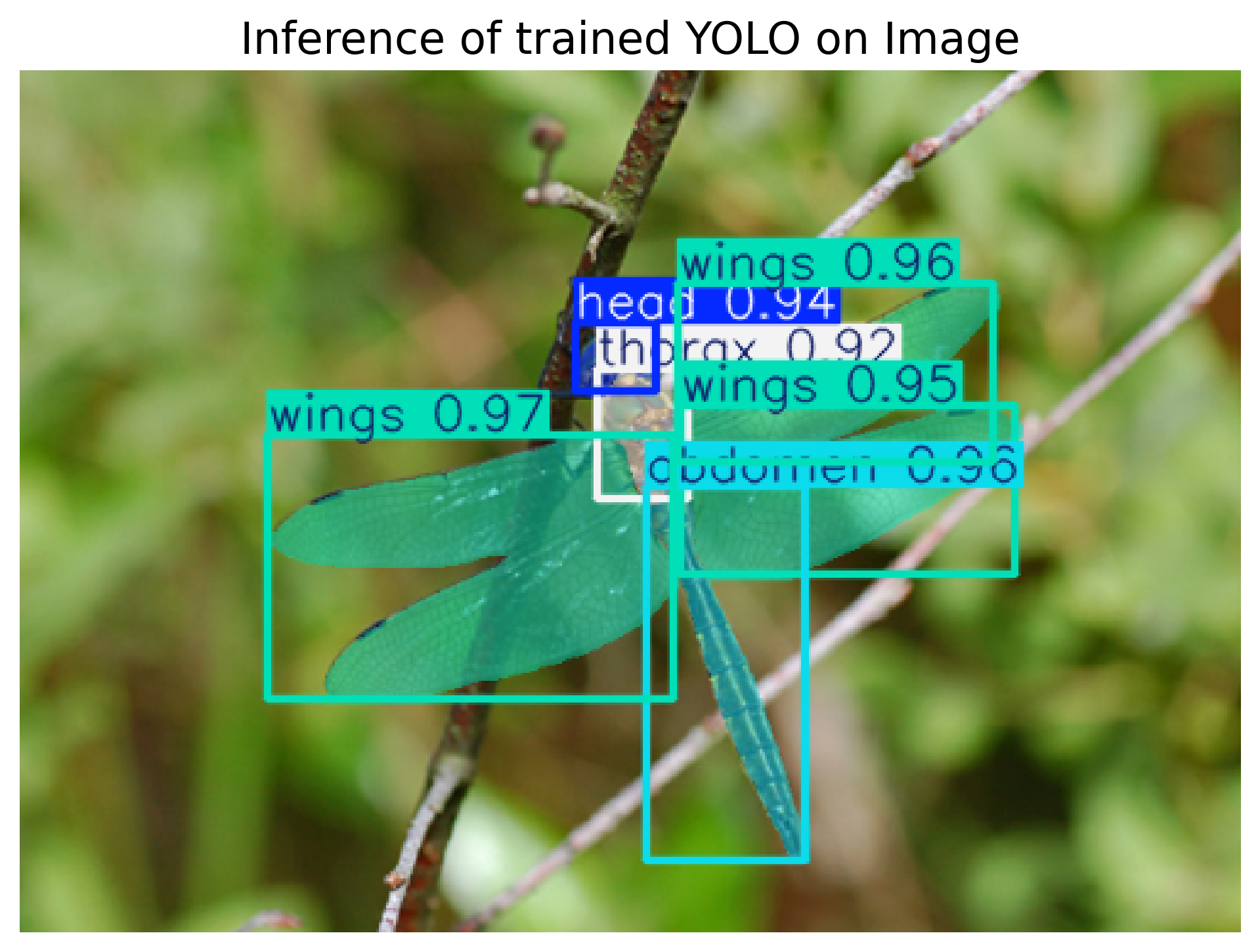}
        \caption{Fine-tuning Stage 2: Results of YOLO-exp02}
        \label{fig:finetune-yolo-2}
    \end{subfigure}
    \caption{Inference of trained YOLO on an unseen image after two rounds of fine-tuning on the dataset.}
    \label{fig:finetuning-inference-results}
\end{figure}

As seen from \autoref{tab:refined-seg-results-bbox} and \autoref{tab:refined-seg-results-seg}, YOLO-exp02 outperformed the other three models. Compared to the first experiment, the mAP scores improved by nearly 10-15\%. Looking at the inference generated by YOLO-exp02 in \autoref{fig:finetune-yolo-2}, we notice that the confidence scores improved slightly for the thorax and abdomen, and it was able to identify individual wing structures. 
As for the other three models, there is a marked improvement in Mask R-CNN and MaskDINO. The performance of Mask R-CNN after 150 epochs exceeded the initial performance of the model after 600 epochs. MaskDINO shows considerable improvement in both classification and segmentation when compared to the initial performance of the model at 100 epochs. Mask2Former still struggles with the classification of the Odonate, but is able to produce partial masks for the parts of the Odonate. 
\begin{table}[H]
\centering
\normalsize   
\begin{tabular}{ccccccccc}
\toprule
    \multirow{2}{*}{\textbf{Model}} & \multirow{2}{*}{\textbf{Epochs}} & \multicolumn{3}{c}{\textbf{Mean Values}} & \multicolumn{4}{c}{\textbf{Per-Class Values}} \\\cmidrule(lr){3-5} \cmidrule(lr){6-9} 
    & & \textbf{mAP} & \textbf{mAP50} & \textbf{mAP75} & \textbf{AP-head}  & \textbf{AP-thorax} & \textbf{AP-Abdomen} & \textbf{AP-wings}  \\ \midrule
    \rowcolor{lightgray} \textbf{YOLO-exp02} & \textbf{150} & \textbf{50.721} & \textbf{88.948} & \textbf{53.264} & \textbf{51.686} & \textbf{44.622} & \textbf{51.303}  & \textbf{55.272}\\ \midrule
    Mask R-CNN & 150  & 15.106 & 39.238 & 9.165 & 23.823 & 25.606 & 10.997 & 0 \\ \midrule
    Mask DINO & 150 & 0.823 & 1.435 & 0.711 & 0.813 &0.813 &0.183 & 2.298 \\ \midrule
    Mask2Former & 150 & 0.018 & 0.079 & 0 & 0.027 & 0 & 0.047& 0 \\\bottomrule
\end{tabular}
\caption{Performance of the model is documented for the \textbf{masks} i.e., segmentation of the \textit{Odonate}, as part of Stage 2 experiments. Training and testing is done on the second version of the dataset \cite{odonata-v2}, as discussed in \autoref{sec:fine-tuning-exp-02}. The metrics are recorded for the mean average precisions and average precision for each class, and the best performing model is in bold. }
\label{tab:refined-seg-results-seg}
\end{table}

\subsection{Results on Colour Extraction}
The results of colour extraction are present in two sections: \autoref{sec:kmeans-results} deals with the results for the extraction of dominant hue, while \autoref{sec:corr-analysis} contains the results for the colour analysis. 
\subsubsection{K-Means Clustering} \label{sec:kmeans-results}
As explained earlier in \autoref{sec:kmeans-methods}, \autoref{alg:kmeans-hue} is used to obtain the final results and the colour palette for each part. 

\begin{figure}[H]
    \centering
    \includegraphics[width=0.7\linewidth]{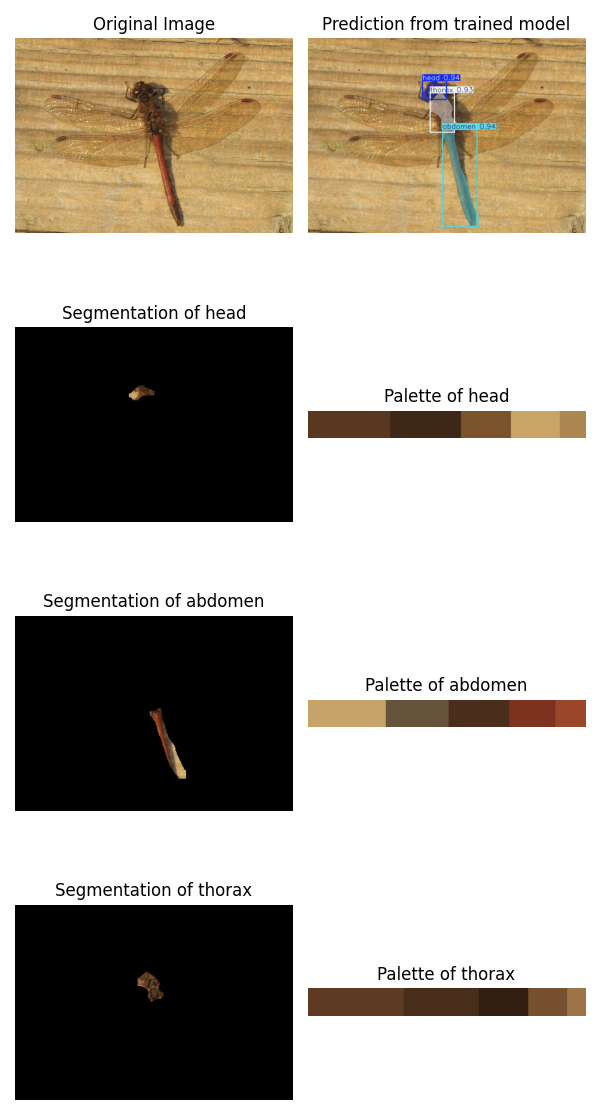}
    \caption{Extraction of colour and dominant hues using K-Means Clustering. The final resultant image is a combination of multiple panels: the original image and the prediction from the model. The other panels contain each identified part and the palette of dominant hues.}
    \label{fig:kmeans-clustering}
\end{figure}

The best performing model \textbf{YOLO-exp02} is used for identification of the parts and generation of masks. The final result is a panel of images, as observed in \autoref{fig:kmeans-clustering}. The first two images are the original and the inference image. The subsequent panels contain the prediction results and the palette constructed from said part. The palette is ordered by occurrences, with the dominant hue at the first, and the width indicating the level of occurrence. 

\subsubsection{Correlation analysis} \label{sec:corr-analysis}
The second experiment on colour analysis establishes a correlation between the average lightness value and the location, as well as the hour of day \footnote{The hour was obtained from the timestamp of the image. The hours were remapped in the following fashion. 20-23 were remapped to 0-3, and 0-19 were remapped to 4-24. This was done to maintain the cycles of day and night. Hour 20 was chosen, as an analysis of the observations showed that most of the images were collected during August and September, and 8 PM was the average sunset time during those months. }. 

The results are grouped according to gender and then by part of the body. The results of only the abdomen are included in this paper, as this was the part with the greatest variation. The results and graphs for other parts are included in \autoref{sec:colour-other}. Pearson and Spearman correlation was done to observe the strength and direction of correlation between the two considered variables.  

Based on the analysis done, it is observed that both the location and the hour of capture have a slight negative correlation with the mean lightness of the body part. This is also evidenced by the value of the correlation coefficient. 

\begin{table}[H]
    \centering
    \begin{tabular}{cccccc}
        \toprule
        \multirow{2}{*}{\textbf{Correlation}} & \multirow{2}{*}{\textbf{Part of the Body}} & \multicolumn{2}{c}{\textbf{Pearson Correlation}} & \multicolumn{2}{c}{\textbf{Spearman Correlation}} \\ \cmidrule(lr){3-4} \cmidrule(lr){5-6}
                 & & \textbf{Correlation} & \textbf{p-value} & \textbf{$\rho$-value} & \textbf{p-value} \\\midrule
            Against Latitude & Abdomen & -0.06283 &  $8.63263 \times 10^{-8}$ & -0.07440 & $2.27792 \times 10^{-10}$\\
            Against hour & Abdomen& -0.04352 &  $2.10521 \times 10^{-4}$& -0.06573& $2.13462 \times 10^{-8}$\\\bottomrule
    \end{tabular}
    \caption{Correlation analysis of mean lightness values of the abdomen of the dragonfly against the latitude. Both Pearson and Spearman correlation coefficient analyses are performed, and the correlation coefficients are tabulated, along with their respective p-values}
     \label{tab:total-results-colour}
\end{table}

\begin{figure}[H]
    \centering
    \begin{subfigure}[b]{0.48\linewidth}
        \centering
        \includegraphics[width=\linewidth]{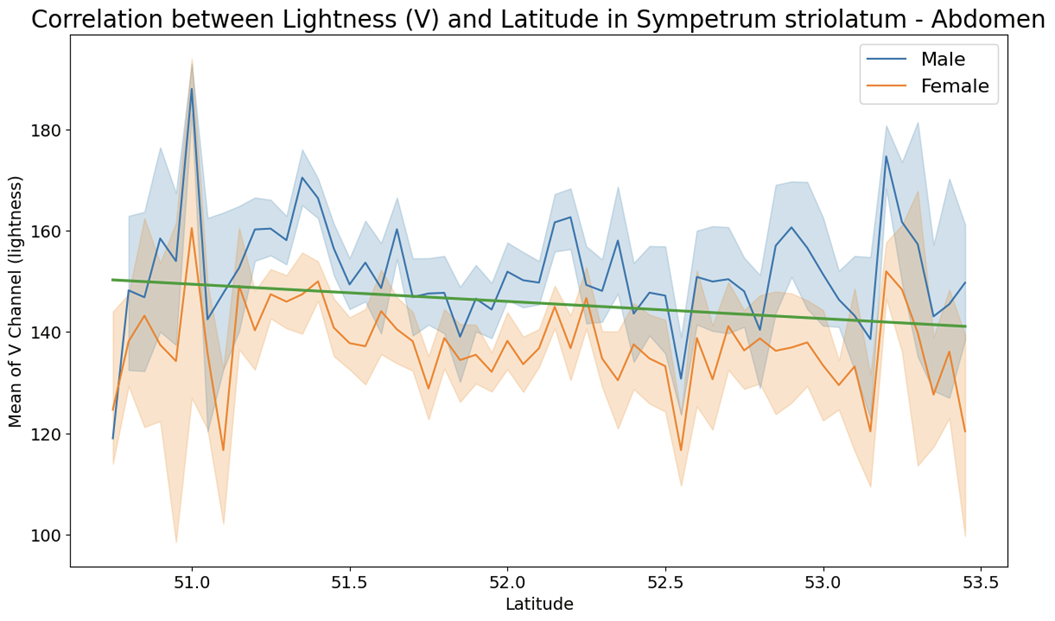}
        \caption{Against latitude}
        \label{fig:lat-abdomen}
    \end{subfigure}
        \begin{subfigure}[b]{0.49\linewidth}
        \centering
        \includegraphics[width=\linewidth]{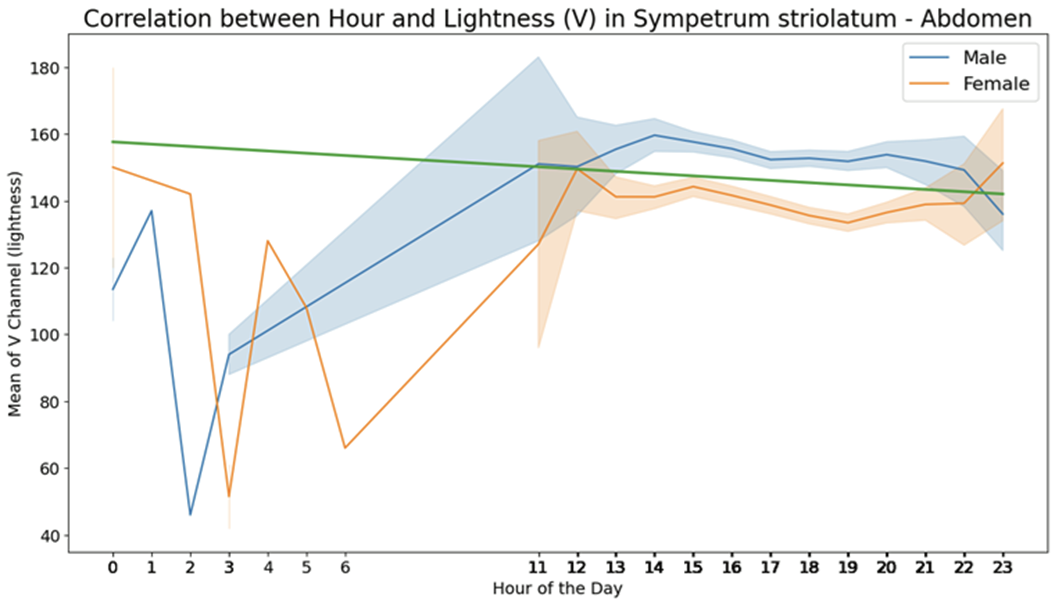}
        \caption{Against the hour of the day}
        \label{fig:hour-abdomen}
    \end{subfigure}
    \caption{Correlation analysis between the lightness (V) of the abdomen and the latitude in \autoref{fig:lat-abdomen}, as well as the hour of day in \autoref{fig:hour-abdomen}. The x-axis corresponds to the latitude, and the hour, while the y-axis corresponds to the lightness of the abdomen. The higher the value, the closer the colour was to white, while the lower the value, the closer the colour was to black. As observed from both graphs, there is a slight negative correlation with the lightness of the abdomen. }
    \label{fig:colour-analysis}
\end{figure}

This shows that the mean lightness is negatively influenced by both the latitude and the hour of the day. As the latitude increases (from Brabant which is $51^\circ$ to Groningen which is $53^\circ$), the part of the body (thorax in this case) gets darker. The same can be said when considering the hour of day. As the natural light and day progresses, the thorax gets darker. 
\section{Conclusion}
In this paper, we implemented a pipeline to recognize Odonates and accurately segment the parts. We prepared a dataset that was collected by citizen science, and available on public platforms (GBIF, in this case). We explored multiple annotation tools and used several to prepare two versions of the dataset that were used for training multiple models. We trained four state-of-the-art models on the two versions of the dataset and measured their training and performance. We performed extensive hyperparameter tuning to arrive at the optimal configuration for the identification and classification of the Odonate. We collaborated with entomologists and computer scientists through surveys to gather insight into the performance and accuracy of the model. 

We propose two methods of extracting colour information: dominant hues using K-Means Clustering, and using the mean values of HSV. We use the lightness (V) values and perform a correlation analysis on the effect of latitude and hour on colouration. In addition to this, we attempt to answer two biological hypotheses on correlation of colour to location and hour of the day. Based on the correlation analyses, we have established that there is a slight weak, negative correlation between the colour and the location, as well as the colour and the hour of day. This paper provides a pipeline for automatic semantic and instance segmentation of Odonates, and a model to identify the parts and extracts colours from the parts.

We believe this work is important, as this allows further analysis on colour information, such as detailed analyses on colour variation based on the seasons. This project also provides a pipeline to extract colour from citizen science data. This would allow expanding on already existing data for other insects, or indigenous species without the need for extensive imaging techniques, or access to historical and archived data. This project also has the potential for application of the same technique on other species or insects. 

\paragraph{Acknowledgements}
This work was performed using the compute resources from the Academic Leiden Interdisciplinary Cluster Environment (ALICE) provided by Leiden University.

\clearpage
\begin{appendix}
\section*{Appendix}
This section consists of additional information on the paper. 
\section{Zero-shot Learning} \label{sec:zero-shot-learning}
Before fine-tuning experiments were performed, an initial zero-shot experiment was done to establish the performance of current models on the dataset.

The models chosen for this task are the same four architectures. However, the models chosen were trained on different datasets. For Mask-R-CNN, MaskDINO and Mask2Former, the final trained models from the paper MassID45 \cite{orsholm_multi-modal_2025} have been used. As for YOLO, there were no models that were pre-trained on insects available. Hence, the official YOLOx11-seg model has been used.

\begin{figure}[H]
    \centering
    \begin{subfigure}[b]{0.4\linewidth}
        \centering
        \includegraphics[width=\linewidth]{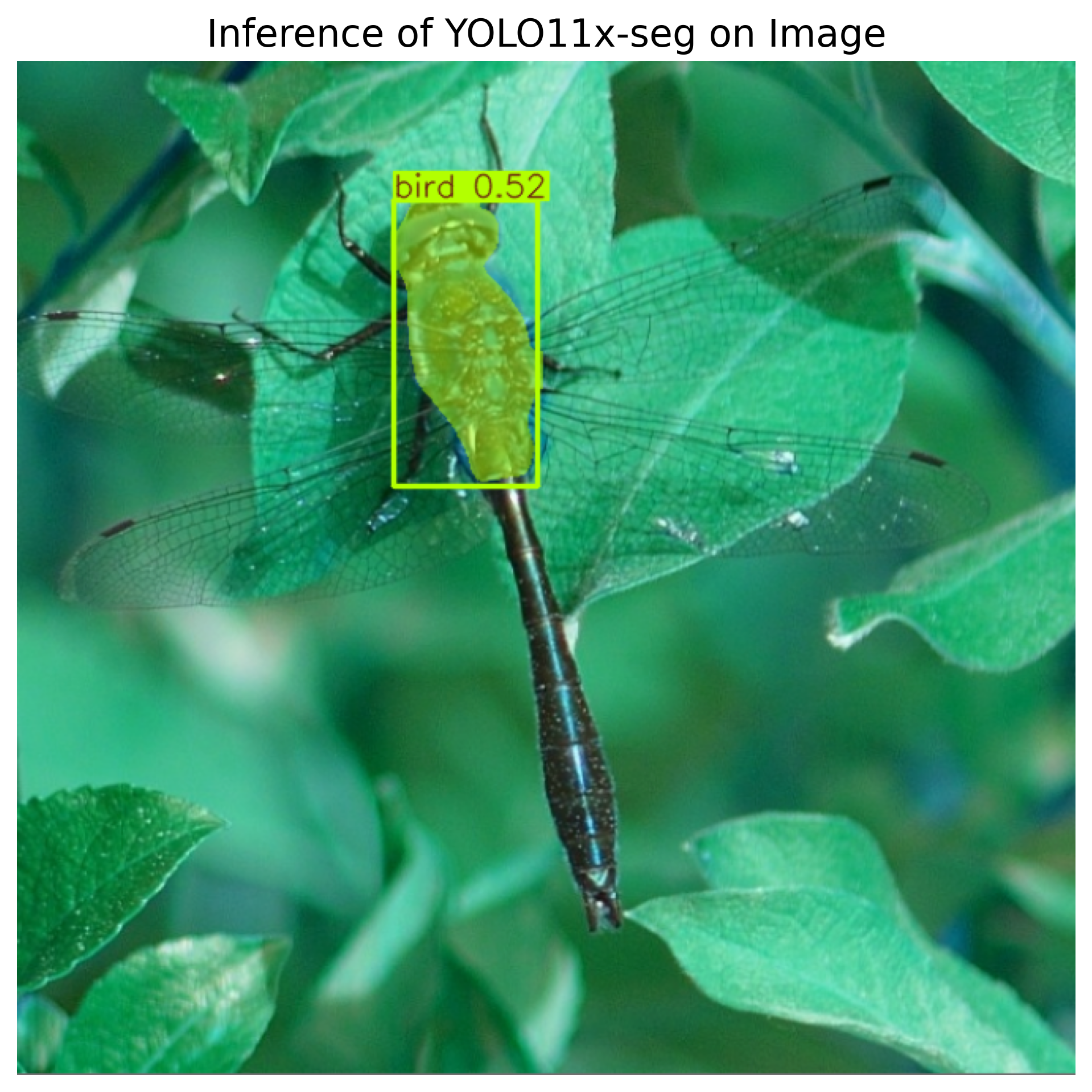}
        \caption{YOLOv11x-seg}
        \label{fig:yolo-zero}
    \end{subfigure}
    \begin{subfigure}[b]{0.4\linewidth}
        \centering
        \includegraphics[width=\linewidth]{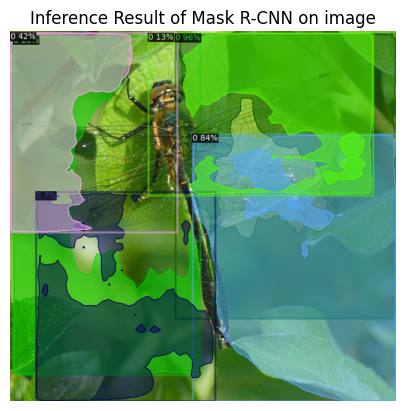}
        \caption{Mask R-CNN}
        \label{fig:rcnn-zero}
    \end{subfigure}
    \begin{subfigure}[b]{0.4\linewidth}
        \centering
        \includegraphics[width=\linewidth]{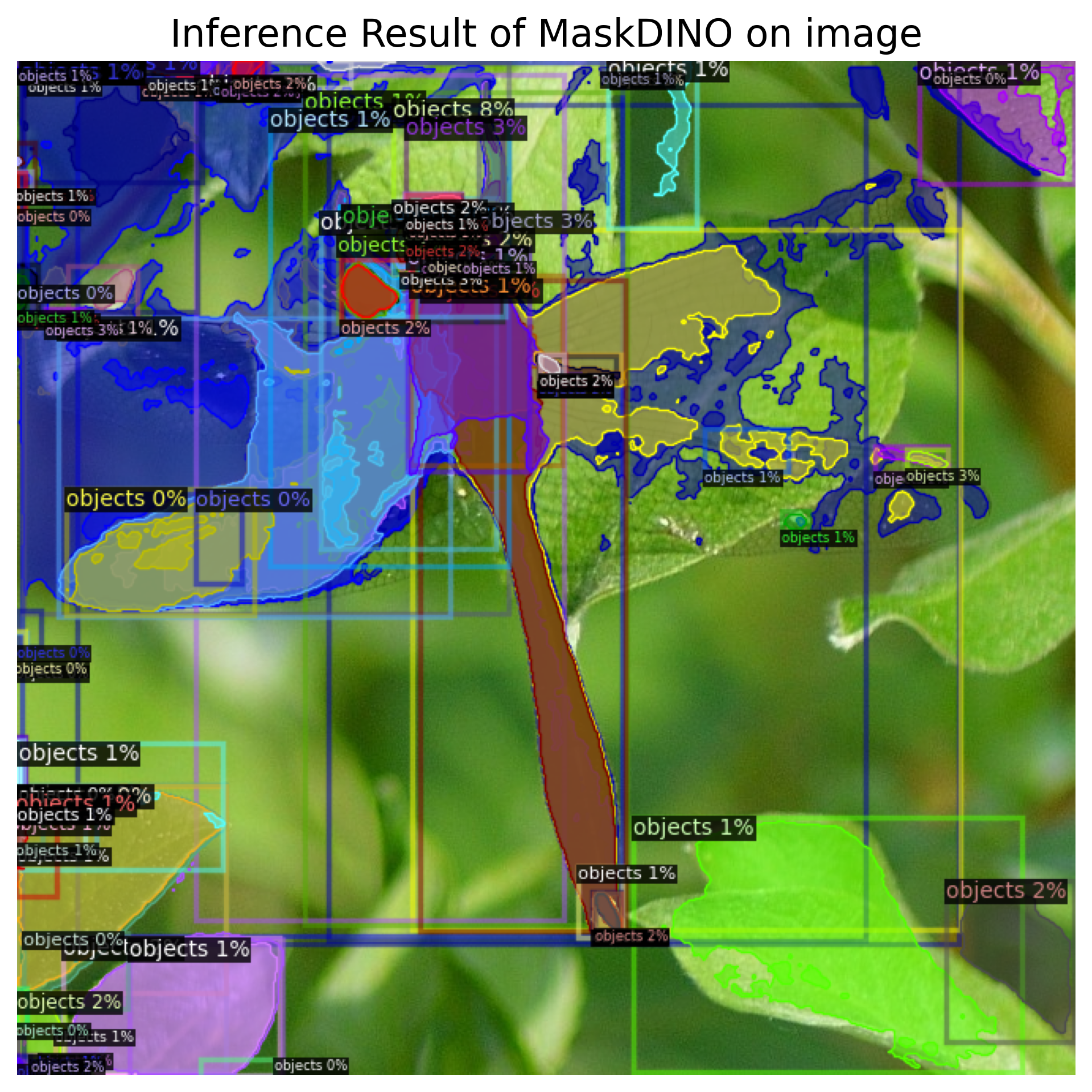}
        \caption{MaskDINO}
        \label{fig:maskdino-zero}
    \end{subfigure}
    \begin{subfigure}[b]{0.4\linewidth}
        \centering
        \includegraphics[width=\linewidth]{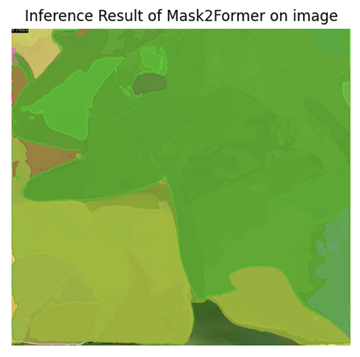}
        \caption{Mask2Former}
        \label{fig:mask2former-zero}
    \end{subfigure}
    \caption{Zero-shot learning on current implementations and models}
    \label{fig:zero-shot-learning}
\end{figure}

Inference was generated on an unseen image from the dataset, and visualized in \autoref{fig:zero-shot-learning}. 

As seen from the images, none of the models were unable to identify or segment the Odonate, much less, the parts. However, this illustrates the need for a model that is trained specifically on Odonates. 

\section{Reliability of Findings} \label{sec:survey-results}
To incorporate the feedback of experts in both fields and to provide an unbiased opinion of the performance of the model after the first fine-tuning experiment, a survey was framed and sent out. The survey has been linked here \cite{dragonfly-formulier-questionnaire}. 

The survey consisted of 150 random images and predictions from YOLO-exp01, and the experts were asked to evaluate the accuracy of the model. The responses were 5 options, as shown in \autoref{tab:survey-response-range}. 

\begin{table}[h!]
\renewcommand{\arraystretch}{1.3}
\centering
\begin{tabularx}{\textwidth}{|c|X|}
\hline
\multirow{3}{*}{Yes} 
 & Yes, only all seven parts (the head, thorax, abdomen, and four wings) are present and recognised by the model. \\ \cline{2-2}
 & Yes, all seven parts are present and recognised by the model. In addition, there are other objects misrecognized by the model. \\ \cline{2-2}
 & Yes, the three main parts (the head, thorax, and abdomen) are recognised, with possible discrepancies in the wings and/or some misrecognised objects. \\
\hline
\multirow{2}{*}{No}
 & No, one or more of the three main parts (the head, thorax, and abdomen) is    not detected in the image. \\ \cline{2-2}
 & No, one of the three main parts (the head, thorax, and abdomen) is misclassified. \\
\hline
\end{tabularx}
\caption{Range of responses in the survey}
\label{tab:survey-response-range}
\end{table}
The responses were structured to give an idea of how well the model performed with semantic segmentation. As the final goal of the paper was to construct a colour palette from the head, thorax and abdomen, two options were provided to see if the model could detect the wings precisely, or just the three required parts. The responses were collected over the span of 3 weeks, and each survey yielded 9 responses. The responses have been tabulated below: \autoref{tab:count-responses-entomologists} contain the responses of the entomologists, and \autoref{tab:count-responses-cs-people} contains the responses of the computer scientists. 
\begin{table}[h!]
\renewcommand{\arraystretch}{1.3} 
\centering 
\begin{tabularx}{\textwidth}{X
                                >{\centering\arraybackslash}c
                                >{\centering\arraybackslash}c
                                >{\centering\arraybackslash}c
                                >{\centering\arraybackslash}c
                                >{\centering\arraybackslash}c
                                >{\centering\arraybackslash}c
                                >{\centering\arraybackslash}c
                                >{\centering\arraybackslash}c
                                >{\centering\arraybackslash}c}
\hline
\multirow{2}{*}{\textbf{Options}} & \multicolumn{9}{c}{\textbf{Respondents}} \\ \cline{2-10}
& 1 & 2 & 3 & 4 & 5 & 6 & 7 & 8 & 9 \\\hline 
No, one of the three main parts (the head, thorax, and abdomen) is misclassified & 26 & 6 & 16 & 9 & 21 & 42 & 9 & 15 & 4 \\ \hline 
No, one or more of the three main parts (the head, thorax, and abdomen) is not detected in the image & 41 & 67 & 54 & 64 & 46 & 45 & 68 & 54 & 70 \\ \hline 
Yes, all seven parts (the head, thorax, abdomen, and four wings) are present and recognised by the model. In addition to this, there are other objects misrecognized by the model. & 9 & 1 & 20 & 2 & 14 & 12 & 10 & 14 & 13 \\ \hline 
Yes, only all seven parts (the head, thorax, abdomen, and four wings) are present and recognised by the model & 37 & 32 & 35 & 41 & 30 & 25 & 27 & 41 & 26 \\ \hline 
Yes, the three main parts (the head, thorax, and abdomen) are recognised, and there can be some discrepancies with the wings, and/or some misrecognised objects & 37 & 43 & 25 & 34 & 39 & 25 & 36 & 26 & 37 \\ \hline 
\end{tabularx} 
\caption{A table containing the count of responses by each entomologist. This has been aggregated over $\approx$ 150 images and omitting missing responses. The 150 images were random, and contained different complexities and background conditions.} 
\label{tab:count-responses-entomologists} 
\end{table}

\begin{table}[h!]
\renewcommand{\arraystretch}{1.3} 
\centering 
\begin{tabularx}{\textwidth}{X
                                >{\centering\arraybackslash}c
                                >{\centering\arraybackslash}c
                                >{\centering\arraybackslash}c
                                >{\centering\arraybackslash}c
                                >{\centering\arraybackslash}c
                                >{\centering\arraybackslash}c
                                >{\centering\arraybackslash}c
                                >{\centering\arraybackslash}c
                                >{\centering\arraybackslash}c}
\hline
\multirow{2}{*}{\textbf{Options}} & \multicolumn{9}{c}{\textbf{Respondents}} \\ \cline{2-10}
& 1 & 2 & 3 & 4 & 5 & 6 & 7 & 8 & 9 \\\hline 
No, one of the three main parts (the head, thorax, and abdomen) is misclassified & 11 & 12 & 24 & 20 & 19 & 18 & 7 & 22 & 10 \\ \hline
No, one or more of the three main parts (the head, thorax, and abdomen) is not detected in the image & 63 & 56 & 51 & 60 & 48 & 57 & 51 & 51 & 62 \\ \hline
Yes, all seven parts (the head, thorax, abdomen, and four wings) are present and recognised by the model. In addition to this, there are other objects misrecognized by the model. & 3 & 9 & 13 & 14 & 22 & 23 & 16 & 1 & 21 \\ \hline
Yes, only all seven parts (the head, thorax, abdomen, and four wings) are present and recognised by the model & 34 & 21 & 26 & 36 & 36 & 30 & 53 & 47 & 37 \\ \hline
Yes, the three main parts (the head, thorax, and abdomen) are recognised, and there can be some discrepancies with the wings, and/or some misrecognised objects & 38 & 52 & 36 & 20 & 25 & 22 & 23 & 29 & 20 \\ \hline
\end{tabularx} 
\caption{A table containing the count of responses by each computer scientist. This has been aggregated over 150 images, of various complexities.} 
\label{tab:count-responses-cs-people} 
\end{table}
Based on the responses, it can be inferred that the responses of the entomologists placed the accuracy of YOLO-exp01 at 17\%-27\% accurate, and the responses of the computer scientists placed the accuracy at 14\%-35\% when the model is asked to recognize the required parts as well as the wings. When only the three main parts are considered, the responses from the entomologists place the accuracy at $\approx$ 28\%, while the response from the computer scientists show that the accuracy stays at 34\%. \footnote{This has been calculated using the upper and lower bound values on the responses for each category. For the first calculation, the responses for the option \textit{Yes, only all seven parts (the head, thorax, abdomen, and four wings) are present and recognised by the model} was considered, and for the second calculation option \textit{Yes, the three main parts (the head, thorax and abdomen) are recognised, with possible discrepancies in the wings and/or some misrecognised objects} was considered. Based on the responses from the entomologists, the count for the first considered option is 25 to 41, and for the second considered option, it is also 25 to 43. Based on the responses from the computer scientists, the count for the first considered option is 21 to 53, whereas for the second considered option, it is 20 to 52. } 

Based on the calculated values, it is established that YOLO-exp01 is fairly accurate, but not accurate enough to carry out the next stage i.e., colour analysis. Hence, this model was used for generating more annotations, and another round of experiments were carried out. 
\section{Colour Analysis} \label{sec:colour-other}
This section contains the results of the colour analysis for both the head and thorax, and the corresponding Spearman and Pearson correlation coefficients. 

\subsection{Against the latitude}
\begin{figure}[h!]
    \centering
    \begin{subfigure}[b]{0.49\linewidth}
        \centering
        \includegraphics[width=\linewidth]{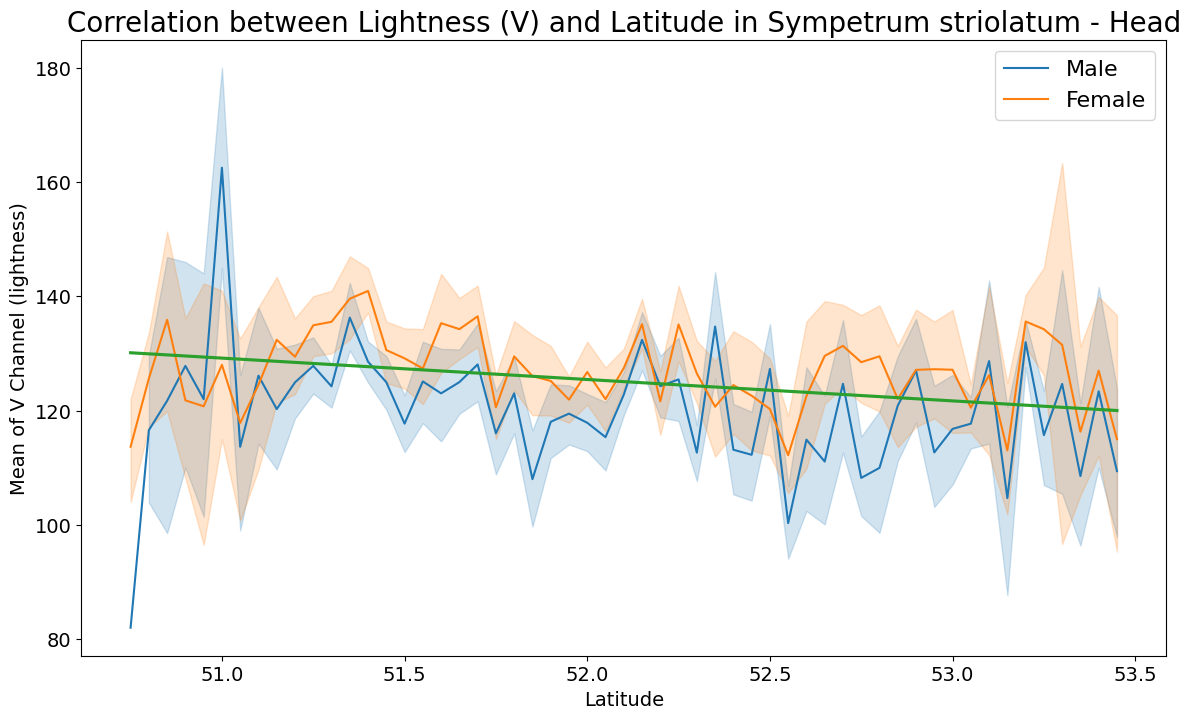}
        \caption{Head}
        \label{fig:latitude-vs-v-head}
    \end{subfigure}
        \begin{subfigure}[b]{0.49\linewidth}
        \centering
        \includegraphics[width=\linewidth]{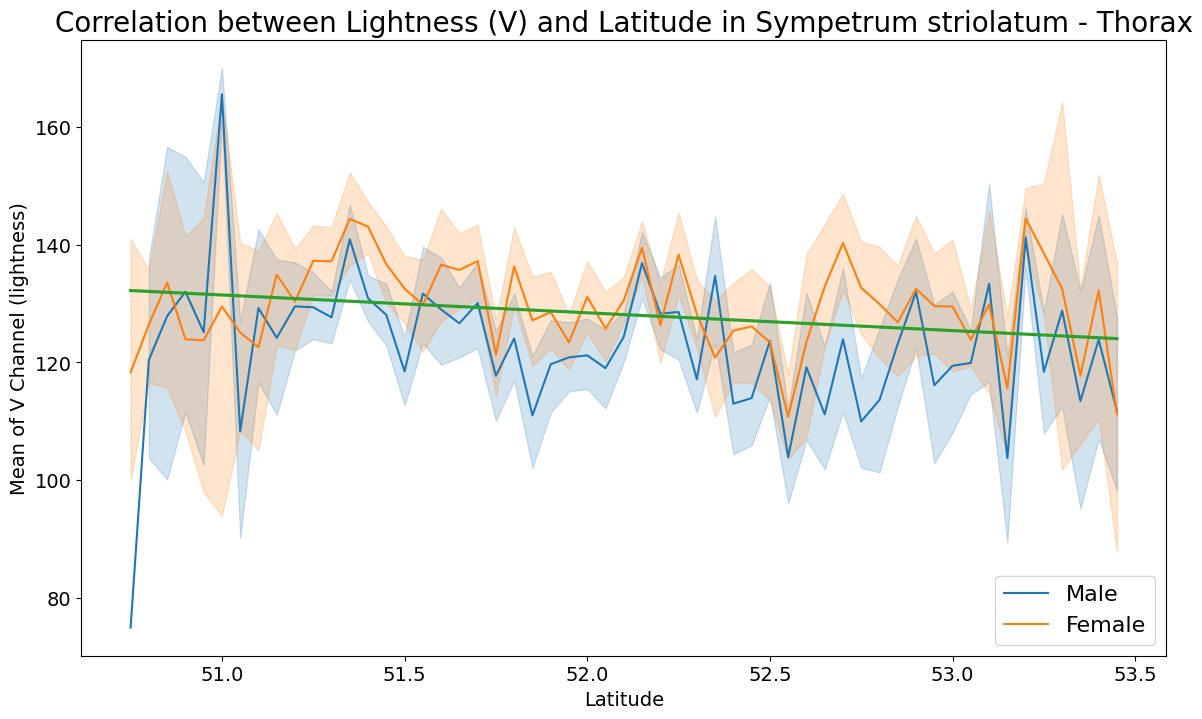}
        \caption{Thorax}
        \label{fig:latitude-vs-v-thorax}
    \end{subfigure}
    \caption{Correlation analysis between the mean lightness (\textbf{V}) values and the latitude. The x-axis corresponds to the latitude, and the y-axis corresponds to the mean lightness value of the body part. The higher the value, the closer the colour was to white, and the lower the value, the closer to black. As observed from both graphs, there is a slight negative correlation between the body part and the lightness}
    \label{fig:correlation-latitude-other-graphs}
\end{figure}

Both head and thorax show a slight negative correlation against the lightness value, as seen in \autoref{tab:latitude-vs-other-dragonfly}. This establishes that as the latitude increases (from Brabant to Groningen), the lightness of the body decreases i.e., it gets darker. This can also be seen from the graphs in \autoref{fig:correlation-latitude-other-graphs}.
\begin{table}[h!]
    \centering
    \begin{tabular}{ccccc}
        \toprule
        \multirow{2}{*}{\textbf{Part of the Body}} & \multicolumn{2}{c}{\textbf{Pearson Correlation}} & \multicolumn{2}{c}{\textbf{Spearman Correlation}} \\ \cmidrule(lr){2-3} \cmidrule(lr){4-5}
        & \textbf{Correlation} & \textbf{p-value} & \textbf{$\rho$-value} & \textbf{p-value} \\\midrule
            Head & -0.07604 & $9.03270 \times 10^{-11}$ & -0.07983 & $1.00323 \times 10^{-11}$ \\ 
            Thorax & -0.05486 & $2.97016 \times 10^{-6}$ &  -0.06111 & $1.92307 \times 10^{-7}$ \\
            \bottomrule
    \end{tabular}
    \caption{Correlation analysis of mean lightness (\textbf{V}) values of the head and thorax of the dragonfly against the latitude. Both Pearson and Spearman correlation coefficient analysis are performed, and the correlation coefficients are tabulated, along with their respective p-values}
    \label{tab:latitude-vs-other-dragonfly}
\end{table}
\subsection{Against the hour of the day}
\begin{figure}[h!]
    \centering
    \begin{subfigure}[b]{0.49\linewidth}
        \centering
        \includegraphics[width=\linewidth]{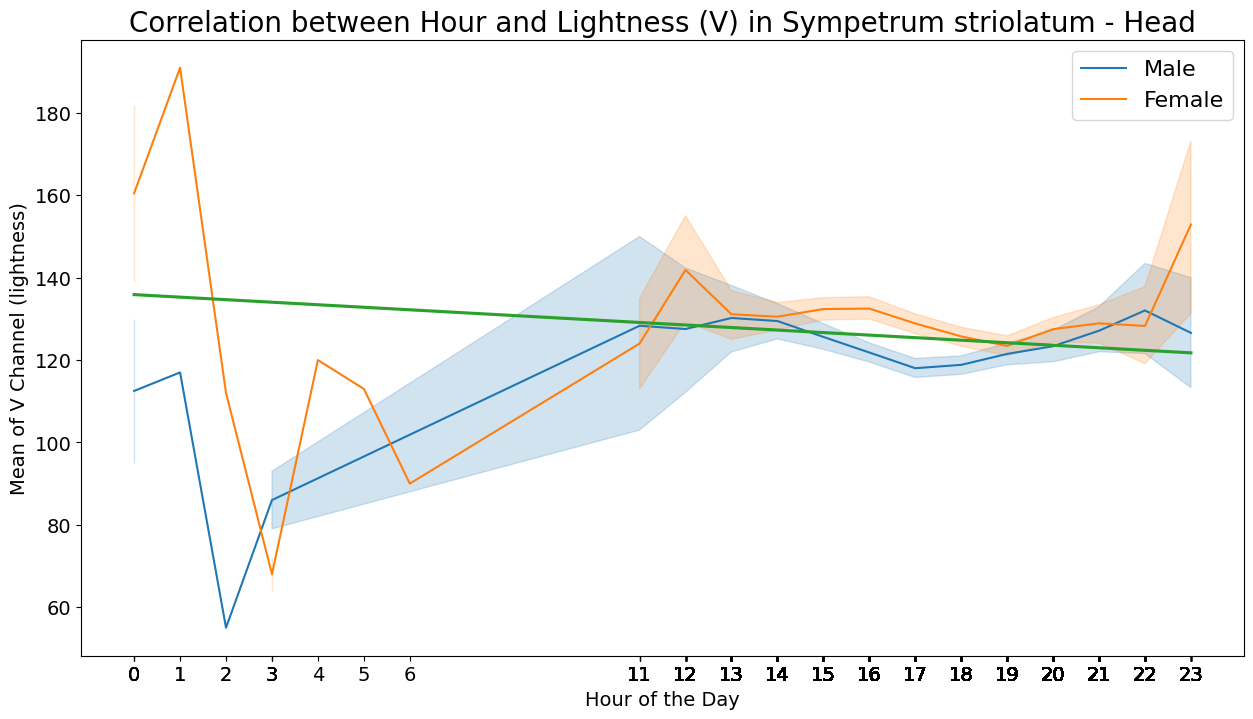}
        \caption{Head}
        \label{fig:hour-vs-v-head}
    \end{subfigure}
        \begin{subfigure}[b]{0.49\linewidth}
        \centering
        \includegraphics[width=\linewidth]{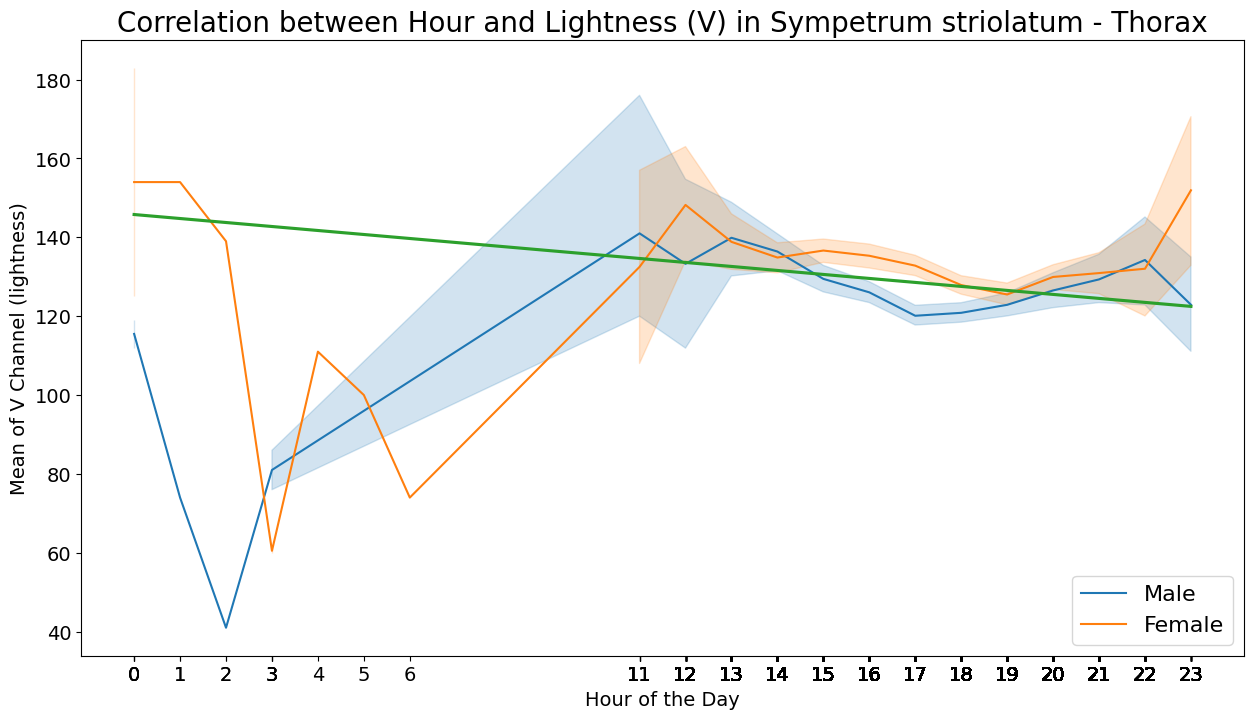}
        \caption{Thorax}
        \label{fig:hour-vs-v-thorax}
    \end{subfigure}
    \caption{Correlation analysis between the mean lightness (\textbf{V}) values and the hour of day. The x-axis corresponds to the adjusted hour of the day, and the y-axis corresponds to the mean lightness value of the body part. The higher the value, the closer the colour was to white, and the lower the value, the closer to black. As observed from both graphs, there is a slight negative correlation between the body part and the lightness}
    \label{fig:correlation-hour-other-graphs}
\end{figure}

Similarly, a correlation analysis was run against the mean lightness (\textbf{V}) values, and the results are provided in \autoref{tab:hours-vs-parts-dragonfly}. This also follows the observed pattern, and has a weak, negative correlation against the hour of the day. This is also evidenced by the graphs, as shown in \autoref{fig:correlation-hour-other-graphs}. 
\begin{table}[H]
    \centering
    \begin{tabular}{ccccc}
        \toprule
        \multirow{2}{*}{\textbf{Part of the Body}} & \multicolumn{2}{c}{\textbf{Pearson Correlation}} & \multicolumn{2}{c}{\textbf{Spearman Correlation}} \\ \cmidrule(lr){2-3} \cmidrule(lr){4-5}
        & \textbf{Correlation} & \textbf{p-value} & \textbf{$\rho$-value} & \textbf{p-value} \\\midrule
            Head&  -0.04341 & $2.18664\times 10^{-4}$ &-0.05433& $3.69492 \times 10^{-6}$\\ 
            Thorax& -0.06394& $5.09281 \times 10^{-8}$&  -0.07939 & $1.30040 \times 10^{-11}$\\\bottomrule
    \end{tabular}
    \caption{Correlation analysis of mean lightness (\textbf{V}) values of the head and thorax of the dragonfly against the hours of the day. Both Pearson and Spearman correlation coefficient analysis are performed, and the correlation coefficients are tabulated, along with their respective p-values}
    \label{tab:hours-vs-parts-dragonfly}
\end{table}
\section{Additional Figures on Inferences Run}
The following sections contain inferences run by the model on an unseen image from the dataset.
\subsection{Fine tuning : Experiment 1}
\begin{figure}[H]
    \centering
    \begin{subfigure}[b]{0.33\linewidth}
        \centering
        \includegraphics[width=\linewidth]{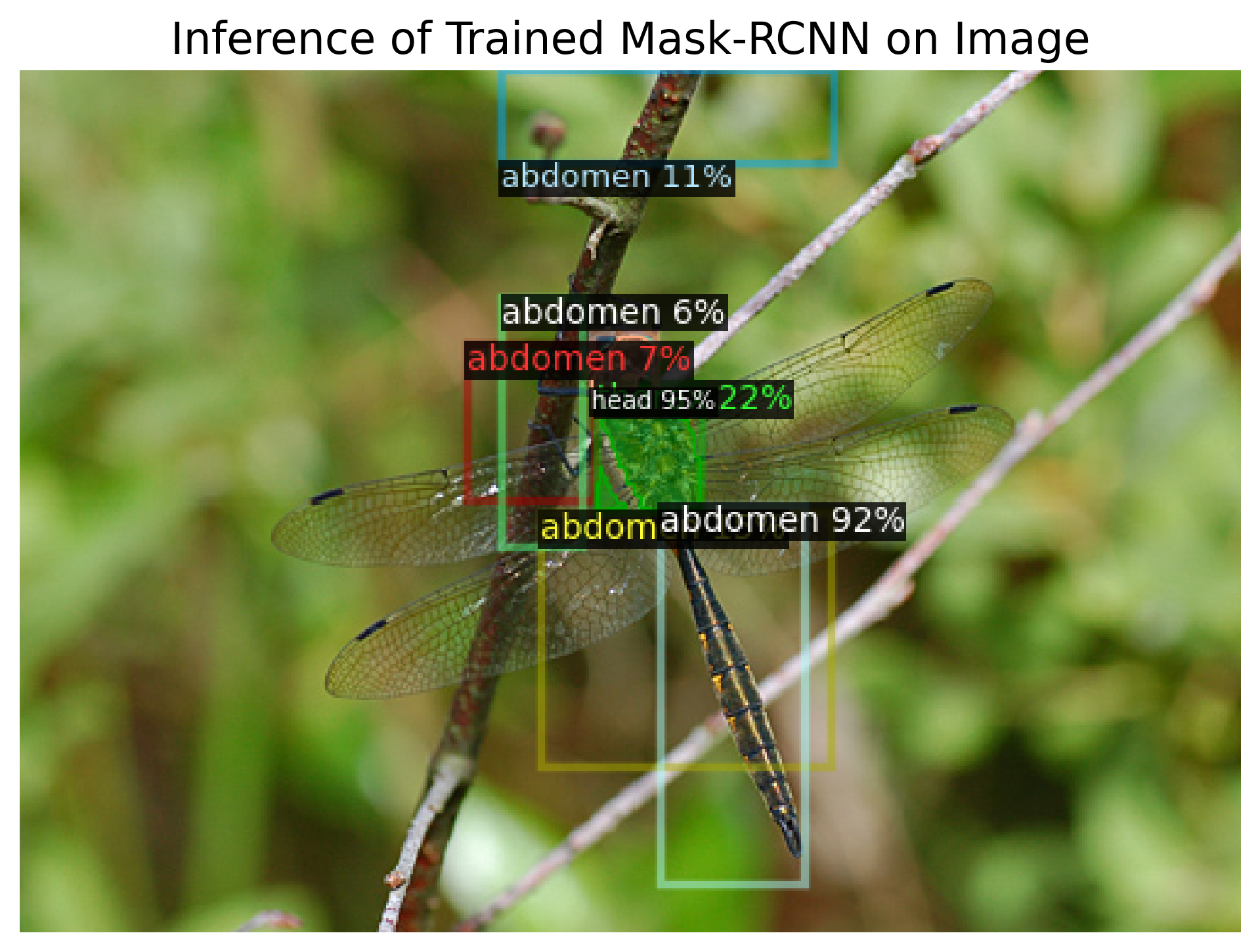}
        \caption{After 100 epochs}
        \label{fig:initial-maskrcnn-2500}
    \end{subfigure}
    \begin{subfigure}[b]{0.33\linewidth}
        \centering
        \includegraphics[width=\linewidth]{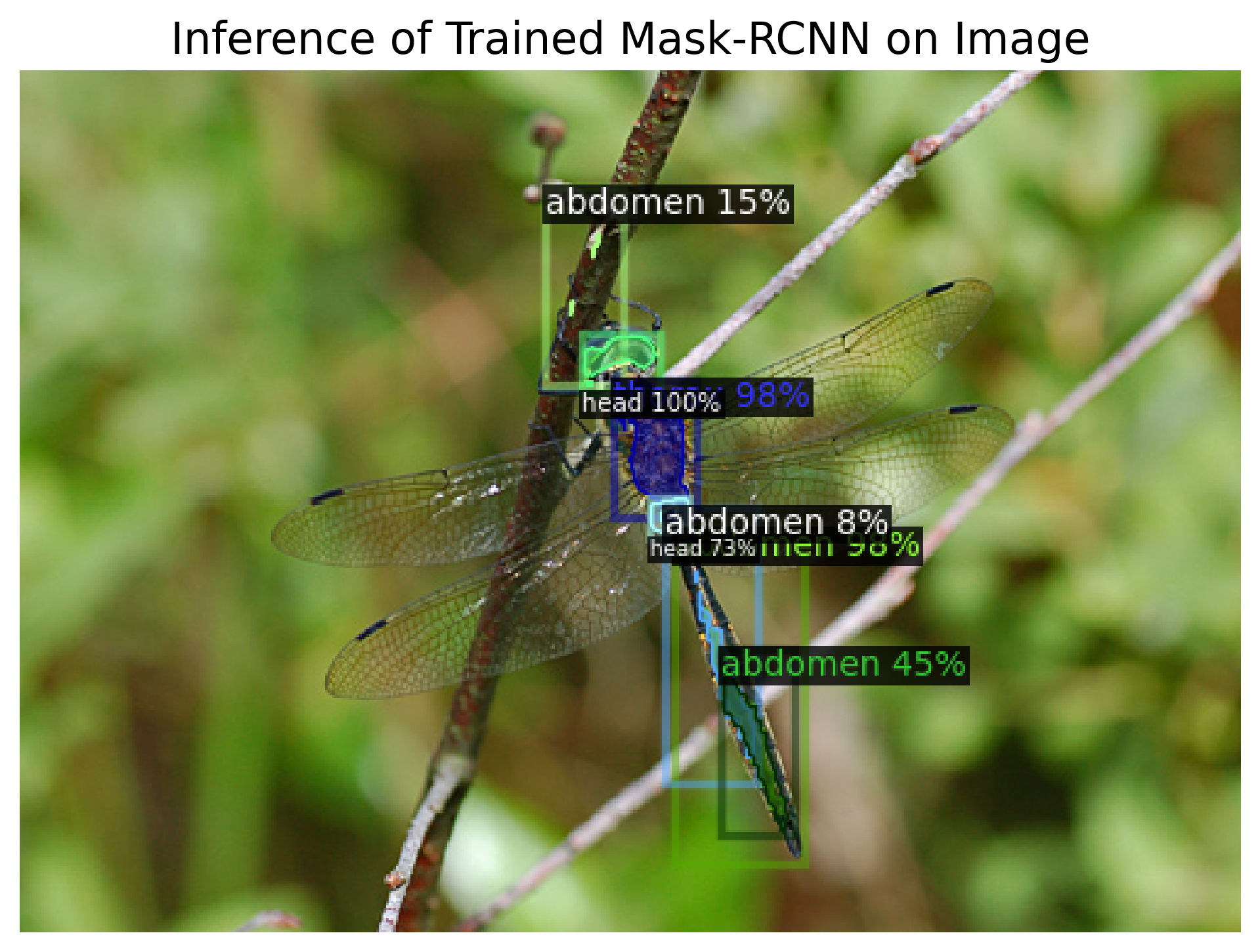}
        \caption{After 200 epochs}
        \label{fig:initial-maskrcnn-5000}
    \end{subfigure}
    \begin{subfigure}[b]{0.33\linewidth}
        \centering
        \includegraphics[width=\linewidth]{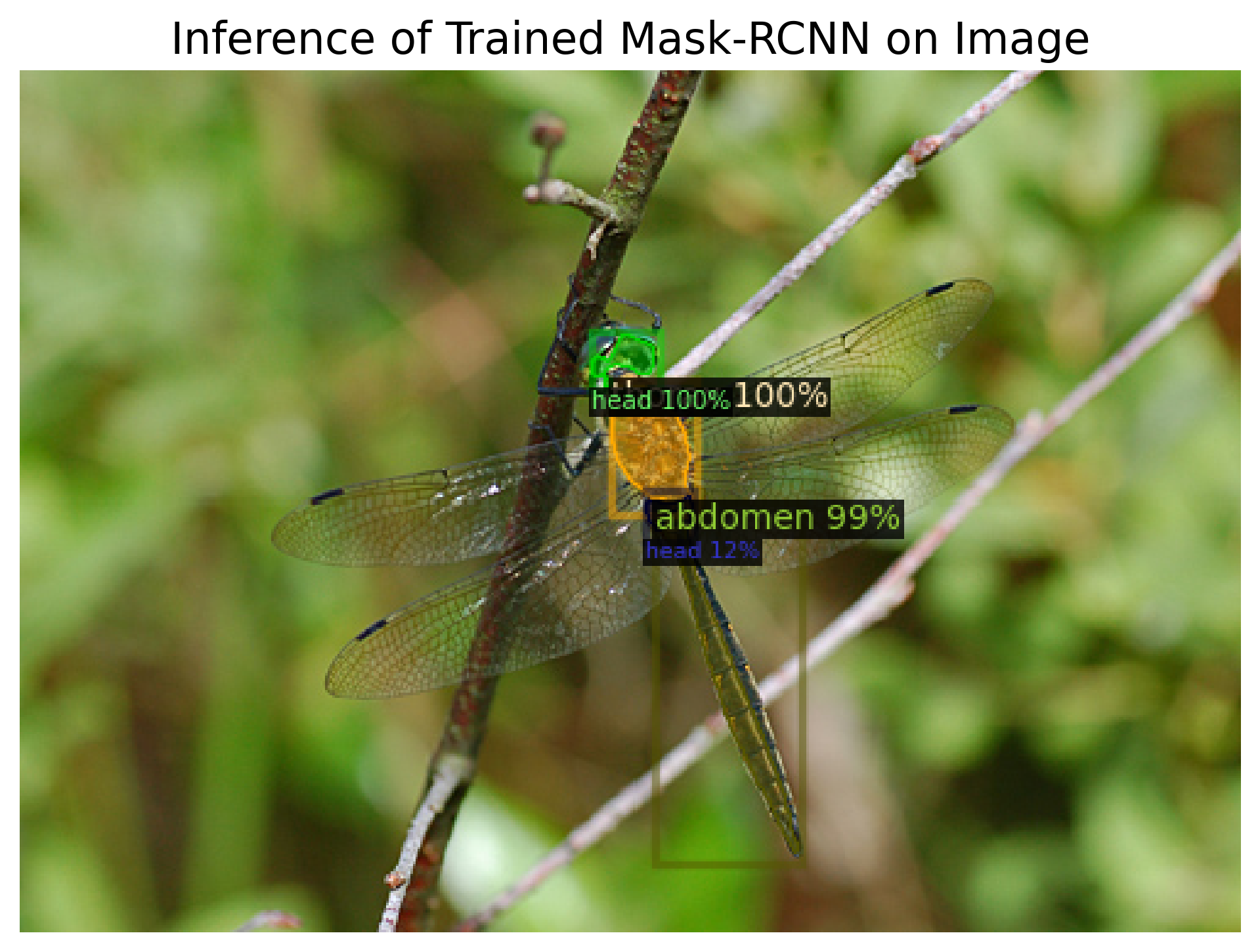}
        \caption{After 600 epochs}
        \label{fig:initial-maskrcnn-15000}
    \end{subfigure}
    \caption{Inferences run by the trained MaskRCNN model at different epochs. }
    \label{fig:initial-runs-maskrcnn}
\end{figure}
\begin{figure}[H]
    \centering
    \begin{subfigure}[b]{0.33\linewidth}
        \centering
        \includegraphics[width=\linewidth]{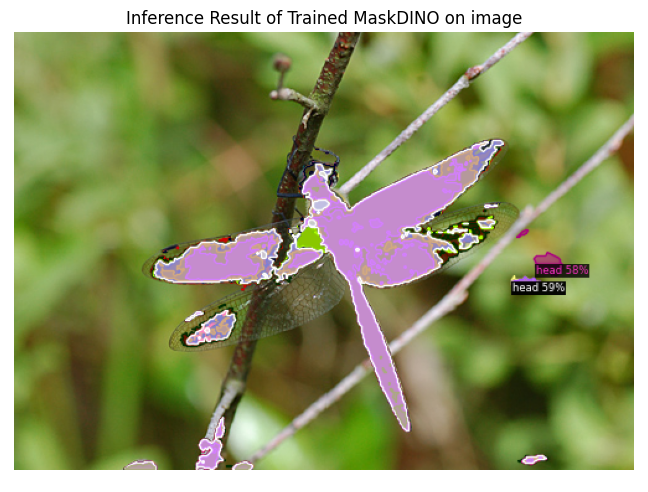}
        \caption{After 100 epochs}
        \label{fig:initial-maskdino-2500}
    \end{subfigure}
    \begin{subfigure}[b]{0.33\linewidth}
        \centering
        \includegraphics[width=\linewidth]{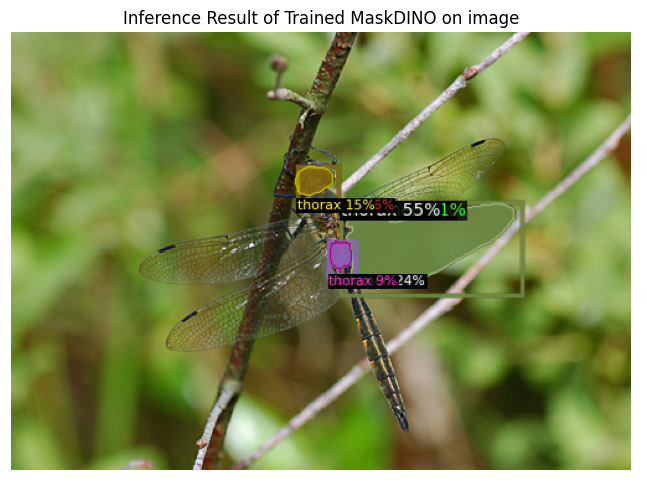}
        \caption{After 200 epochs}
        \label{fig:initial-maskdino-5000}
    \end{subfigure}
    \begin{subfigure}[b]{0.33\linewidth}
        \centering
        \includegraphics[width=\linewidth]{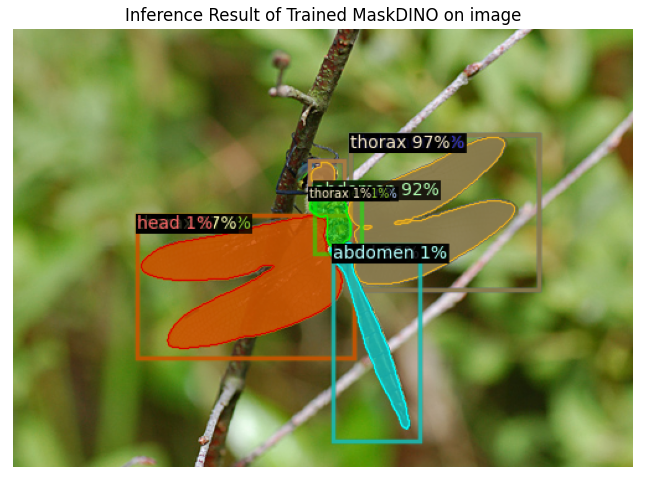}
        \caption{After 600 epochs}
        \label{fig:initial-maskdino-15000}
    \end{subfigure}
    \caption{Inferences run by the trained MaskDINO model at different epochs. }
    \label{fig:initial-runs-maskdino}
\end{figure}
\begin{figure}[H]
    \centering
    \begin{subfigure}[b]{0.33\linewidth}
        \centering
        \includegraphics[width=\linewidth]{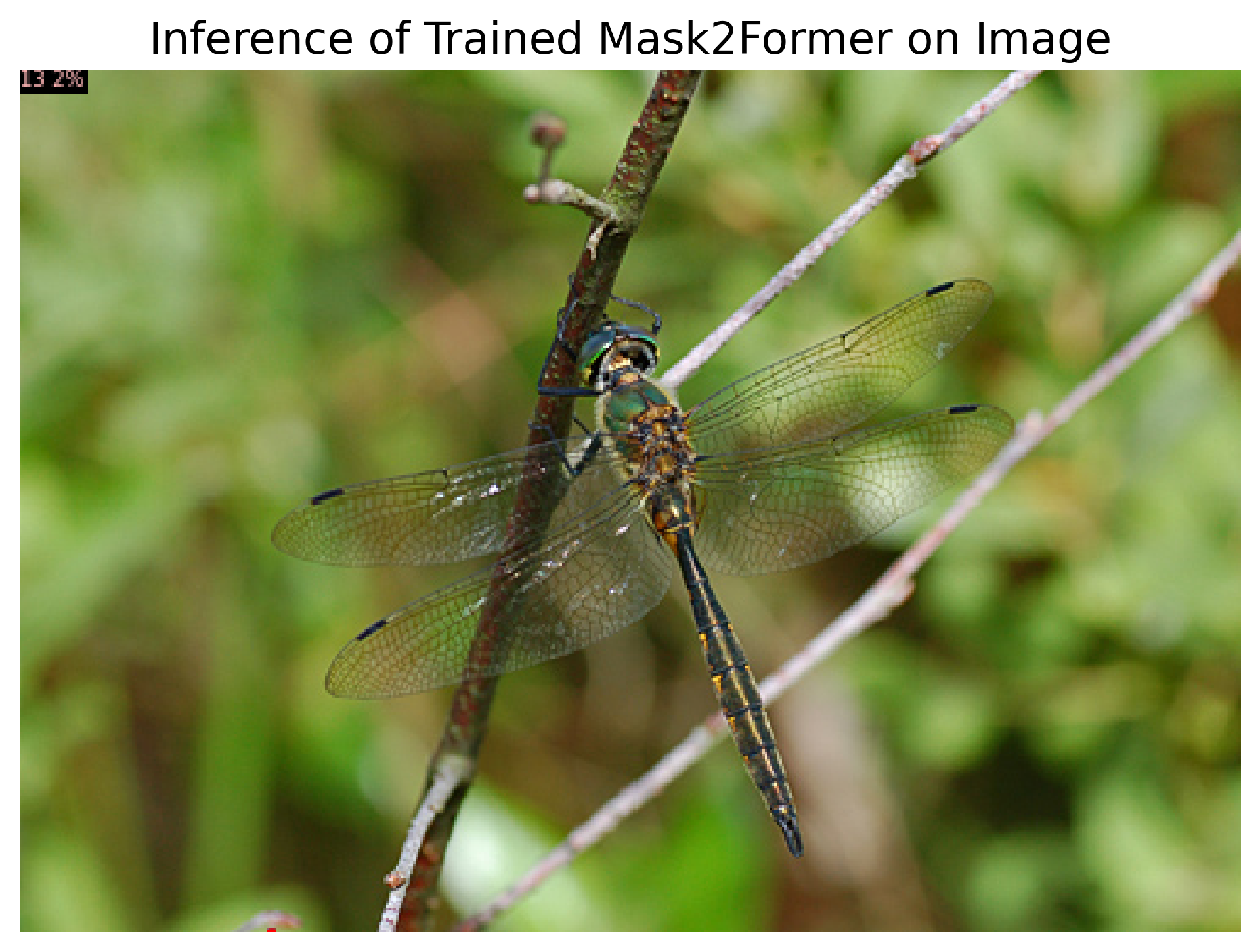}
        \caption{After 100 epochs}
        \label{fig:initial-mask2former-2500}
    \end{subfigure}
    \begin{subfigure}[b]{0.33\linewidth}
        \centering
        \includegraphics[width=\linewidth]{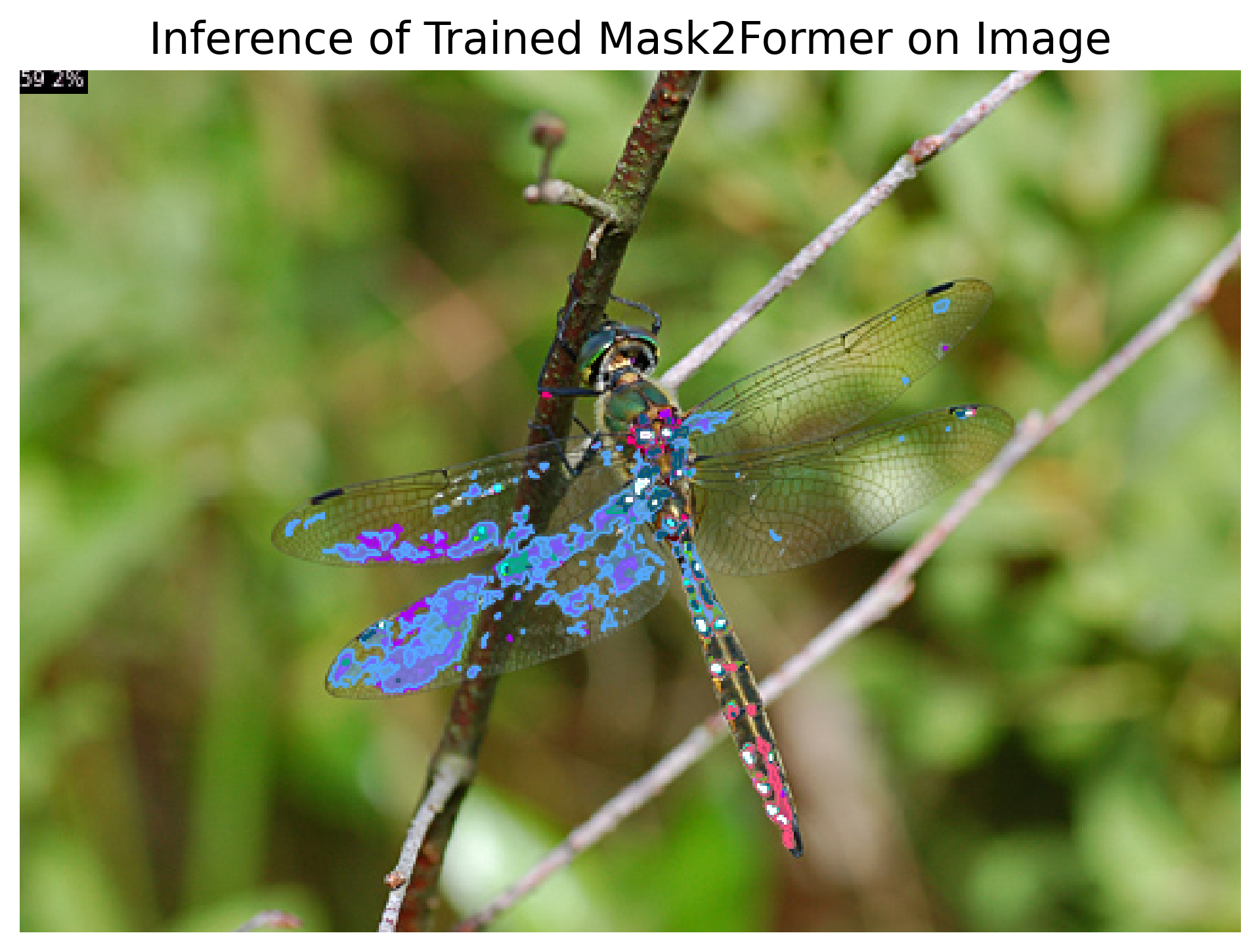}
        \caption{After 200 epochs}
        \label{fig:initial-mask2former-5000}
    \end{subfigure}
    \begin{subfigure}[b]{0.33\linewidth}
        \centering
        \includegraphics[width=\linewidth]{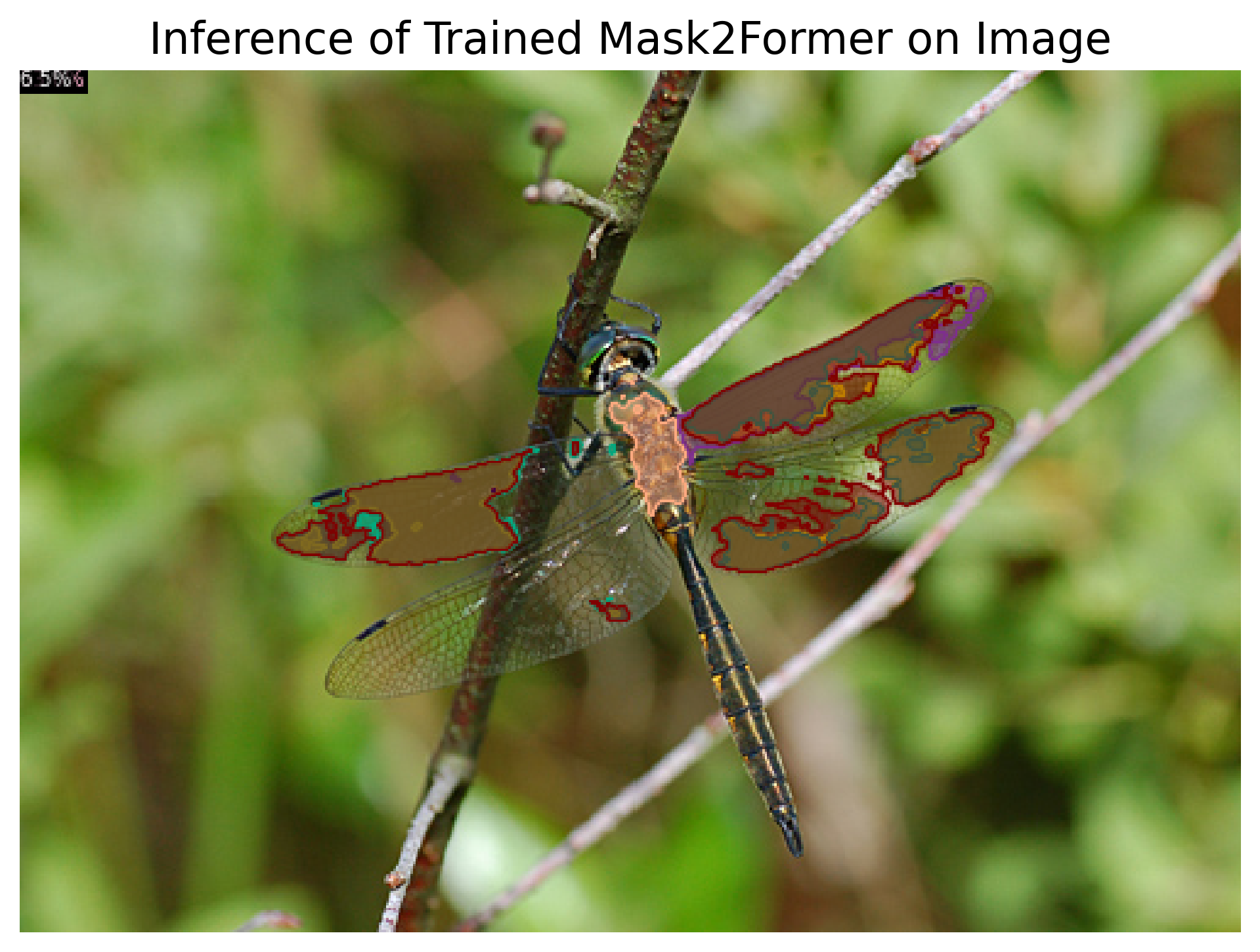}
        \caption{After 600 epochs}
        \label{fig:initial-mask2former-15000}
    \end{subfigure}
    \caption{Inferences run by the trained Mask2Former model at different epochs. }
    \label{fig:initial-runs-mask2former}
\end{figure}
\subsection{Fine Tuning : Experiment 2}
\begin{figure}[H]
    \centering
    \begin{subfigure}[b]{0.33\linewidth}
        \centering
        \includegraphics[width=\linewidth]{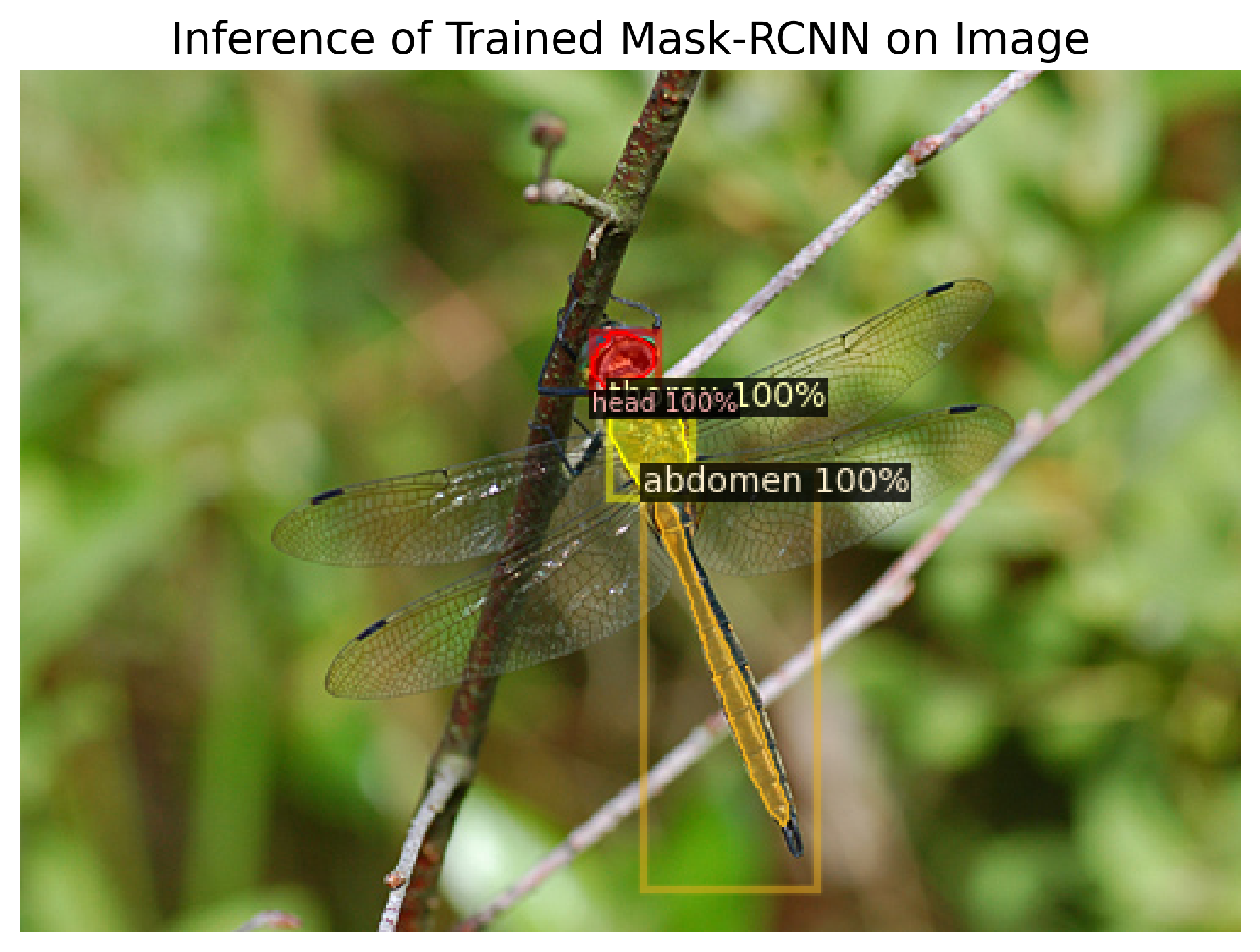}
        \caption{MaskRCNN: After 150 epochs}
        \label{fig:refined-maskrcnn-150}
    \end{subfigure}
    \begin{subfigure}[b]{0.33\linewidth}
        \centering
        \includegraphics[width=\linewidth]{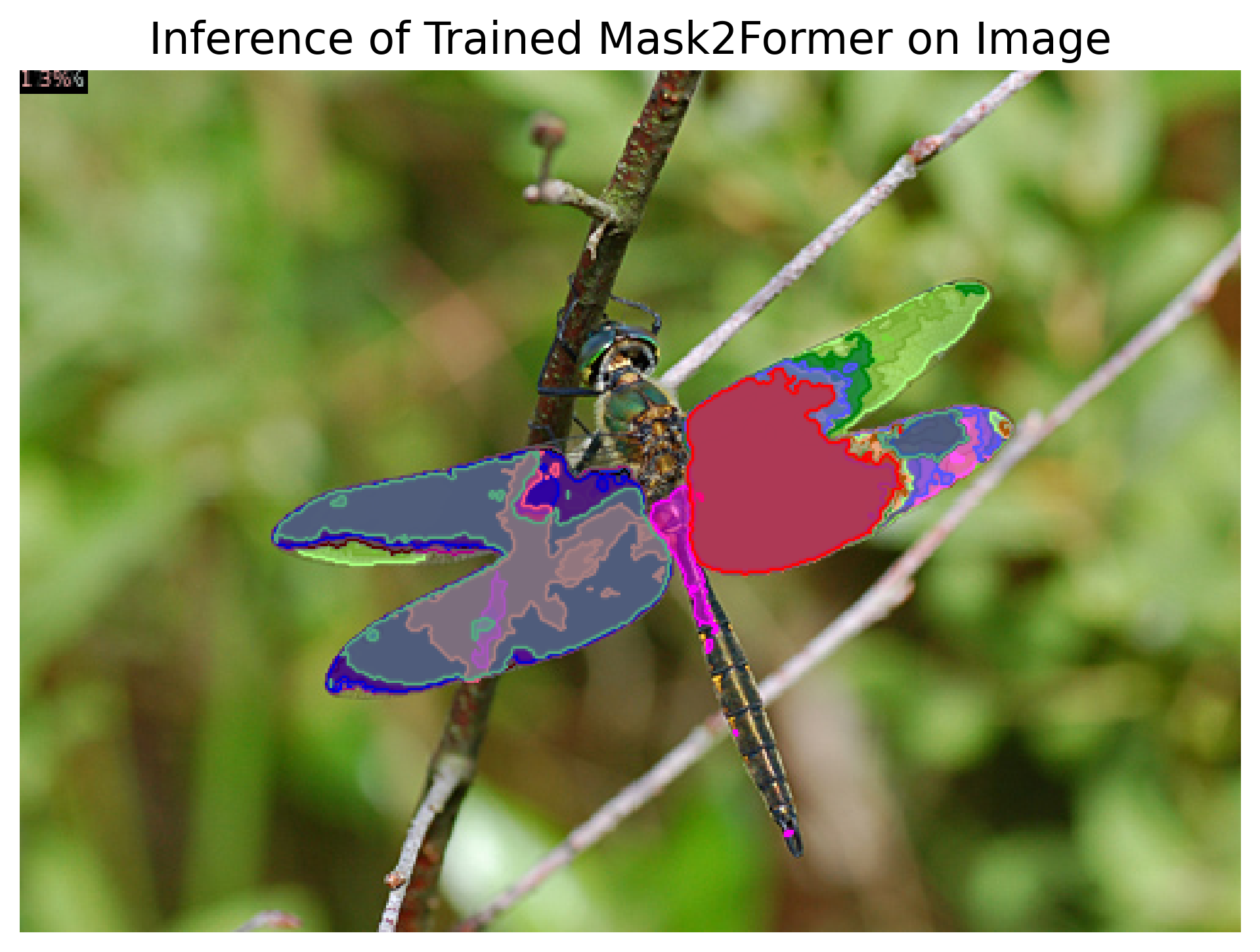}
        \caption{Mask2Former: After 150 epochs}
        \label{fig:refined-mask2former-150}
    \end{subfigure}
    \begin{subfigure}[b]{0.33\linewidth}
        \centering
        \includegraphics[width=\linewidth]{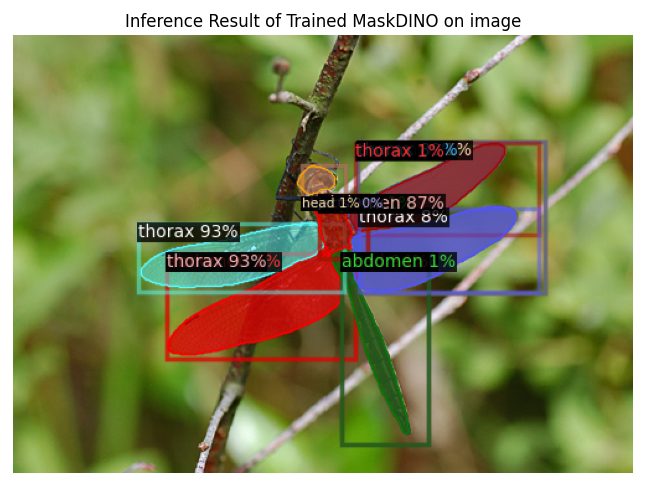}
        \caption{MaskDINO: After 150 epochs}
        \label{fig:refined-maskdino-150}
    \end{subfigure}
    \caption{Inferences run by the trained models on the second version of the dataset, at 150 epochs}
    \label{fig:refined-runs-all}
\end{figure}
\end{appendix}
\clearpage

\bibliographystyle{unsrt}
\bibliography{references} 
\end{document}